\documentclass[11pt]{article}

\usepackage[final]{acl}
\usepackage{algorithm}
\usepackage{algpseudocode}
\usepackage{amsmath,amssymb}
\usepackage{array}
\usepackage{booktabs}
\usepackage{capt-of}
\usepackage{cuted}
\usepackage{enumitem}
\usepackage[T1]{fontenc}
\usepackage{graphicx}
\usepackage[utf8]{inputenc}
\usepackage{inconsolata}
\usepackage{latexsym}
\usepackage{mathtools}
\usepackage{pgfplotstable}
\usepackage{placeins}
\usepackage{tabularx}
\usepackage{tikz}
\usetikzlibrary{positioning}
\usepackage[most]{tcolorbox}
\usepackage{times}
\usepackage{xcolor}
\pgfplotsset{compat=1.18}

\allowdisplaybreaks

\title{Beyond Meta-Reasoning: Metacognitive Consolidation for Self-Improving LLM Reasoning}

\author{
  Ziqing Zhuang\textsuperscript{1,\textdagger},
  Linhai Zhang\textsuperscript{2,\textdagger},
  Jiasheng Si\textsuperscript{4},
  Deyu Zhou\textsuperscript{1,*},
  Yulan He\textsuperscript{2,3} \\
  \textsuperscript{1}Southeast University \quad
  \textsuperscript{2}King's College London \\
  \textsuperscript{3}The Alan Turing Institute \quad
  \textsuperscript{4}Qilu University of Technology (Shandong Academy of Sciences) \\
  \texttt{\{ziqingzhuang, d.zhou\}@seu.edu.cn} \\
  \texttt{\{linhai.zhang, yulan.he\}@kcl.ac.uk} \\
  \texttt{jiashengsi@qlu.edu.cn}
}

\begin{document}
\maketitle
\begingroup
\renewcommand{\thefootnote}{\fnsymbol{footnote}}
\footnotetext[2]{These authors contributed equally.}
\footnotetext[1]{Corresponding author.}
\endgroup
\setcounter{footnote}{0}
\begin{abstract}
Large language models (LLMs) have demonstrated strong reasoning capabilities, and as existing approaches for enhancing LLM reasoning continue to mature, increasing attention has shifted toward meta-reasoning as a promising direction for further improvement.
However, most existing meta-reasoning methods remain episodic: they focus on executing complex meta-reasoning routines within individual instances, but ignore the accumulation of reusable meta-reasoning skills across instances, leading to recurring failure modes and repeatedly high metacognitive effort.
In this paper, we introduce Metacognitive Consolidation, a novel framework in which a model consolidates metacognitive experience from past reasoning episodes into reusable knowledge that improves future meta-reasoning.
We instantiate this framework by structuring instance-level problem solving into distinct roles for reasoning, monitoring, and control to generate rich, attributable meta-level traces. These traces are then consolidated through a hierarchical, multi-timescale update mechanism that gradually forms evolving meta-knowledge.
Experimental results demonstrate consistent performance gains across benchmarks and backbone models, and show that performance improves as metacognitive experience accumulates over time.
\end{abstract}

\section{Introduction}

Large Language Models (LLMs) have demonstrated strong reasoning capabilities, and extensive work has further advanced them through reinforcement learning~\cite{openai2024openaio1card,guo2025deepseek} and test-time scaling~\cite{snell2025scaling,wu2025inference}.
As these traditional routes mature, attention is increasingly shifting to meta-reasoning, the ability to reason about how to reason, which offers a complementary perspective for improving the intelligence of LLMs~\cite{wang-zhao-2024-metacognitive,yan2025position}.

\begin{figure}[t]
    \centering
    \includegraphics[width=0.45\textwidth]{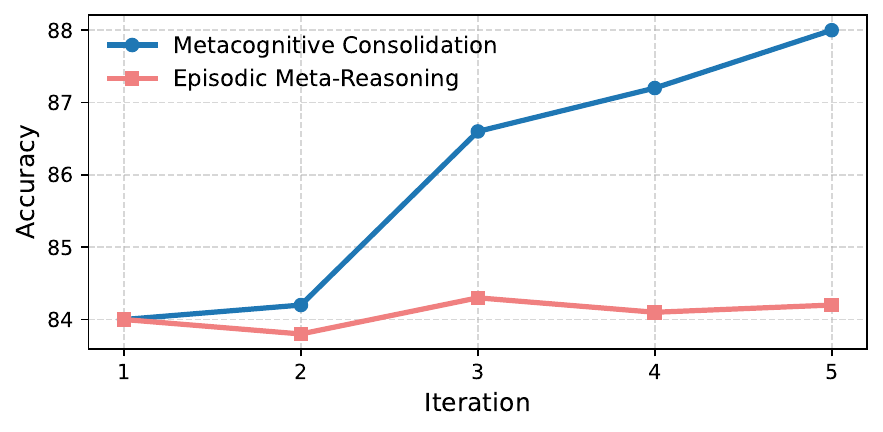}
    \caption{\textbf{Metacognitive Consolidation} goes beyond episodic meta-reasoning by accumulating experience across instances, leading to steadily improving performance as test-time experience grows (\textcolor{blue}{blue}), whereas traditional meta-reasoning remains episodic and exhibits limited or fluctuating gains (\textcolor{red}{red}). (benchmark: MATH-500)}
    \label{fig:intro}
\end{figure}

Existing research on LLM meta-reasoning largely follows two paradigms.
The first is pre-task meta-planning, where the model generates guiding meta-information, such as a reasoning plan, strategy, or scaffolding, before executing the actual reasoning~\cite{gao2024meta,suzgun2024meta,wan2025rema}.
The second is in-task meta-control, where the model enters reasoning directly but injects meta-level interventions during the trajectory to steer, revise, or re-plan intermediate steps~\cite{zhang2025rlvmr,yang2025test,xu2025thinker}.

However, most existing meta-reasoning methods remain episodic.
They focus solely on the execution of meta-reasoning acts within a single instance, ignoring the acquisition of meta-reasoning skills over time.
As a result, similar failure modes recur on recurring problems, and the system repeatedly relies on the same heavy meta-reasoning routines (e.g., retries, verification, or intervention) to recover~\cite{tan-etal-2025-consistent}.
This lack of skill carryover prevents amortizing metacognitive effort over time, leaving the model largely \textit{cognitively static} rather than evolving into a progressively more robust reasoner~\cite{he-etal-2025-large}.

To address this challenge, we introduce \textbf{Metacognitive Consolidation}.
Beyond performing meta-reasoning, a model should consolidate meta-level experience accumulated across reasoning episodes into reusable procedural knowledge that improves future meta-reasoning.
Theoretically, this echoes the cognitive account of \textit{proceduralization}~\cite{anderson1982acquisition}, where skills transition from an explicit, declarative stage to a procedural stage in which the same competence can be executed more directly with practice.
For LLMs, this mechanism is pivotal as it converts episodic test-time compute into persistent capability growth, enabling the model to evolve from a static solver into a self-improving reasoner.

To enable metacognitive consolidation, we require structured and attributable meta-level feedback about what failed and how it was corrected, whereas standard Chain-of-Thought~\cite{wei2022chain} often entangles such signals into an opaque, monolithic stream.
Accordingly, we propose \textbf{Meta-Reasoning Orchestrator} (MRO), a modular architecture in which three specialized agents collaboratively augment the reasoning process with explicit meta-level structure.
The \textit{Monitor} audits the \textit{Reasoner}'s trajectory and the \textit{Controller} intervenes based on this feedback, consistent with cognitive accounts of meta-reasoning as monitoring and control~\cite{ACKERMAN2017607}.
This factorization yields fine-grained action--critique--correction traces that support metacognitive consolidation while also improving inference-time reliability through explicit oversight and targeted intervention.

To realize metacognitive consolidation, we introduce \textbf{MetaCognitive Accumulator} (MCA), a hierarchical memory module that retains and updates meta-level experience across reasoning instances at multiple temporal frequencies.
The MCA aggregates \textit{instance-level reflections} into \textit{batch-level micro-lessons}, and further integrates them into \textit{long-term meta-knowledge} that evolves more slowly over time.
This hierarchical accumulation enables continual adaptation while balancing retention and forgetting, reflecting the role of bounded, multi-timescale memory in modern reasoning systems~\cite{behrouz2025nested,behrouz2025titans}.

Together, MRO and MCA form an inner--outer loop architecture, named as MC$^2$ (MetaCognitive Consolidation), where MRO performs instance-level meta-reasoning and MCA accumulates and updates meta-knowledge across instances.
To demonstrate the effectiveness of our method, we evaluate it across multiple backbone models and reasoning benchmarks, showing consistent improvements over strong baselines.
Moreover, as shown in Figure~\ref{fig:intro}, these improvements grow with experience, indicating that the model progressively accumulates and leverages metacognitive knowledge over time.

Our contributions are threefold:
\begin{itemize}
    \item \textbf{New Framework}: We introduce Metacognitive Consolidation, a new perspective that extends meta-reasoning beyond episodic, instance-level control to the accumulation of reusable metacognitive skills.
    \item \textbf{Framework Instantiation}: We propose MC$^2$, an inner--outer loop framework consisting of a Meta-Reasoning Orchestrator (MRO) and a MetaCognitive Accumulator (MCA), enabling structured meta-reasoning within instances and hierarchical consolidation across instances.
    \item \textbf{Empirical Performance}: Through extensive experiments across multiple models and reasoning benchmarks, we show that metacognitive consolidation yields consistent gains that grow with experience.
\end{itemize}

\section{Method}
\label{sec:method}

\begin{figure*}[t]
    \centering
    \includegraphics[width=0.9\textwidth]{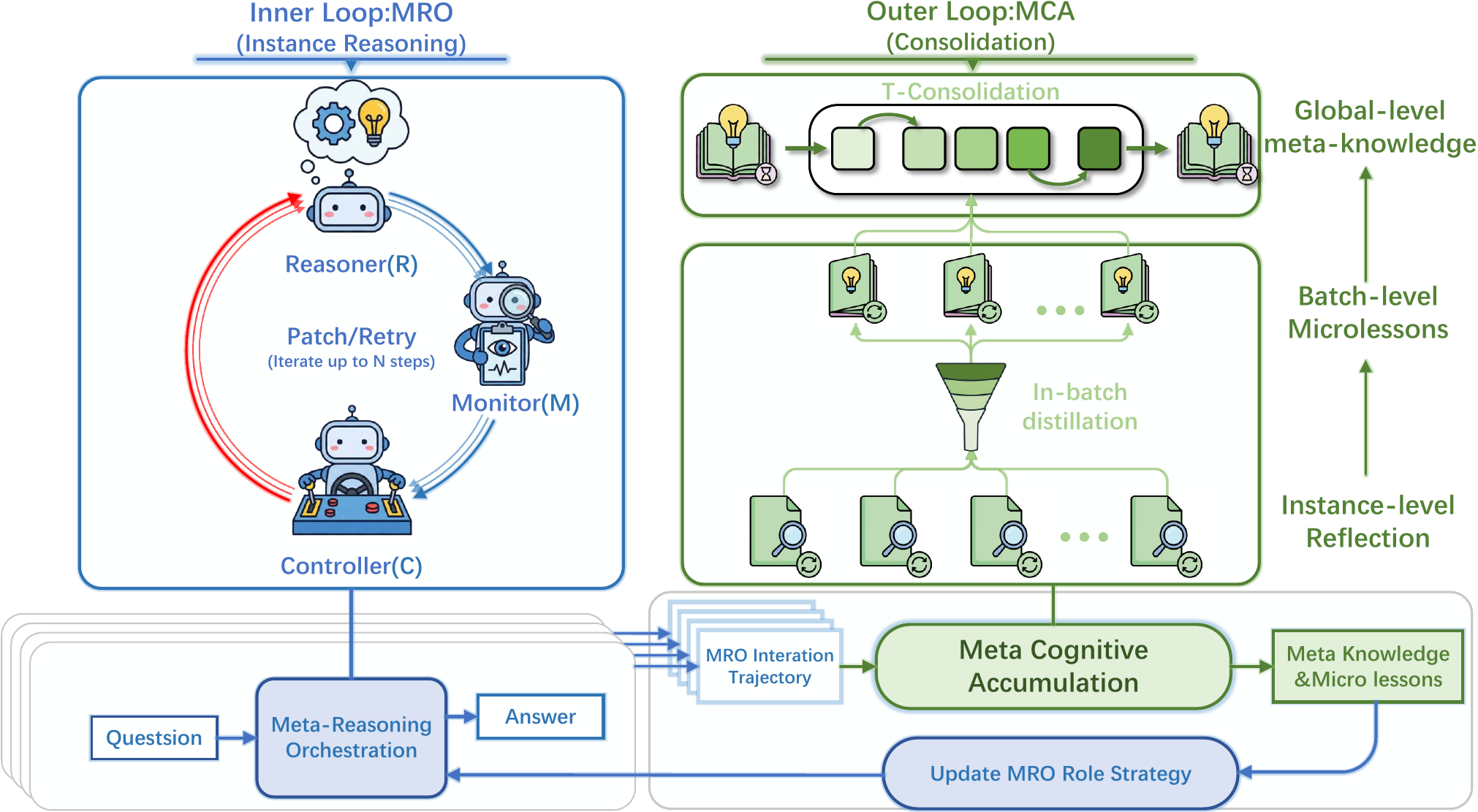}
    \caption{Overview of \textbf{Metacognitive Consolidation (MC$^2$)}.
    The inner loop (MRO) produces structured action--critique--correction traces via a Reasoner--Monitor--Controller decomposition, while the outer loop (MCA) consolidates these traces across instances into evolving meta-knowledge and updates role-specific policies for subsequent inference.}
    \label{fig:method}
\end{figure*}

We consider a reasoning stream $\mathcal{D}=\{(x_t,y_t^\star)\}_{t=1}^{T}$, where $x_t$ is an input instance and $y_t^\star$ is the gold answer.
Let $\mathrm{LLM}_\theta(\cdot)$ denote a frozen backbone model.
Given a prompt, the model generates an answer $\hat{y}_t$ together with a reasoning trajectory $\tau_t$ (e.g., a chain-of-thought).

\paragraph{Episodic reasoning.}
In the standard (episodic) setting, each instance is solved independently with a fixed inference prompt $P$:
\begin{equation}
\label{eq:standard}
(\hat{y}_t,\tau_t)=\mathrm{LLM}_\theta(x_t\mid P).
\end{equation}

\paragraph{Test-time evolvement.}
In contrast, under \emph{test-time evolvement}, the inference policy is allowed to \emph{update online} as the system processes more instances.
We model this by maintaining a time-indexed policy $P_t$ that is updated using experience from previous reasoning trajectories:
\begin{equation}
\label{eq:policy_update_general}
P_{t+1} = \mathrm{update}(P_t, \tau_t),
\end{equation}
and then used to solve the next instance:
\begin{equation}
\label{eq:evolve_general}
(\hat{y}_{t+1},\tau_{t+1})=\mathrm{LLM}_\theta(x_{t+1}\mid P_{t+1}).
\end{equation}

\paragraph{Overall framework.}
As shown in Figure~\ref{fig:method}, MC$^2$ consists of two coupled modules.
\textbf{Meta-Reasoning Orchestrator (MRO)} is the inner loop that performs instance-level meta-reasoning by coordinating a \textit{Reasoner}, \textit{Monitor}, and \textit{Controller}, producing both the final answer and an explicit action--critique--correction trace.
\textbf{MetaCognitive Accumulator (MCA)} is the outer loop that consolidates these traces across instances into meta-knowledge at multiple temporal frequencies and uses it to form stronger role policies over time.

\subsection{Meta-Reasoning Orchestrator}
\label{sec:mro}

To consolidate the metacognitive experience, the system needs structured and attributable signals about \emph{what failed} and \emph{what fixed it}.
In contrast, standard chain-of-thought reasoning~\cite{wei2022chain} primarily exposes the solution trajectory itself, while many self-revision pipelines (e.g., self-refinement or verification) often entangle critique and correction into a single opaque stream~\cite{madaan2023selfrefine}, making it harder to extract reusable meta-level signals.

Motivated by the cognitive science view of meta-reasoning as \emph{monitoring and control of thinking}~\cite{ACKERMAN2017607}, we decompose instance-level inference into three specialized agents Reasoner/Monitor/Controller with role-specific policy prompts $P_t\triangleq(P_{R,t},P_{M,t},P_{C,t})$. Within the MRO, the Reasoner proposes a solution trajectory, the Monitor audits it, and the Controller decides whether to accept, patch, or request a revision.
This interaction is implemented as an iterative loop: the Controller's feedback is passed back to the Reasoner to guide the next attempt, and the loop repeats until a termination condition is met or a maximum of $N$ iterations is reached.

\paragraph{Reasoner.}
At inner iteration $k$, the Reasoner generates a candidate answer and trajectory conditioned on the input and the Controller's previous feedback $u_t^{k-1}$:
\begin{equation}
\label{eq:reasoner}
(\hat{y}_t^{k}, \tau_t^{k})
= \mathcal{R}\!\left(x_t, u_t^{k-1}\,|\,P_{R,t}\right),
\end{equation}
where $u_t^{-1}=\emptyset$ for initialization.
Intuitively, $u_t^{k-1}$ can encode targeted constraints such as ``re-check step $j$'' or ``restart with an alternative approach.''

\paragraph{Monitor.}
Given the Reasoner's proposal, the Monitor audits the trajectory and produces a diagnostic report:
\begin{equation}
\label{eq:monitor}
g_t^{k}
= \mathcal{M}\!\left(x_t, \hat{y}_t^{k}, \tau_t^{k}\,|\,P_{M,t}\right),
\end{equation}
where $g_t^{k}$ typically includes (i) a verdict, (ii) an error localization signal (e.g., the first suspicious step), and (iii) a concise critique or evidence.

\paragraph{Controller.}
The Controller decides whether to accept, patch, or restart based on the candidate and the Monitor's report:
\begin{equation}
\label{eq:controller}
(a_t^{k}, \tilde{y}_t^{k}, u_t^{k})
= \mathcal{C}\!\left(x_t, \hat{y}_t^{k}, \tau_t^{k}, g_t^{k}\,|\,P_{C,t}\right),
\end{equation}
where the action $a_t^{k}\in\{\textsc{Accept},\textsc{Patch},\\\textsc{Restart}\}$,
$\tilde{y}_t^{k}$ is an optional patched answer (only used when $a_t^{k}=\textsc{Patch}$),
and $u_t^{k}$ is feedback passed to the next Reasoner call (only used when $a_t^{k}=\textsc{Restart}$).

The MRO loop terminates when either (i) $a_t^{k}\in\{\textsc{Accept},\textsc{Patch}\}$ or (ii) $k=N-1$ (iteration budget exhausted).
We define the resulting structured trace as:
\begin{equation}
\label{eq:trace}
\mathcal{T}_t
= \Big\{ \big(\hat{y}_t^{k}, \tau_t^{k}, g_t^{k}, a_t^{k}\big) \Big\}_{k=0}^{K_t-1},
\quad K_t\le N.
\end{equation}
This factorization yields fine-grained action--critique--correction records that are directly usable for consolidation, while also improving inference-time robustness via explicit oversight and targeted intervention.

\subsection{MetaCognitive Accumulator}
\label{sec:mca}

MCA consolidates the meta-level experience produced by MRO into reusable knowledge that improves future meta-reasoning.
A key challenge is that raw traces $\{\mathcal{T}_t\}$ are heterogeneous and noisy across instances, and cannot be directly ``stored'' as meta-knowledge without redundancy and drift.
We therefore adopt a \emph{hierarchical, multi-frequency} consolidation scheme with three levels:
\textit{instance-level reflections} $\rightarrow$ \textit{batch-level micro-lessons} $\rightarrow$ \textit{global-level meta-knowledge}.

Since Reasoner, Monitor, and Controller serve distinct roles, MCA processes their experiences \emph{separately} and maintains three role-specific consolidation streams.
As the consolidation pipeline is identical across roles, we omit the $\mathcal{R}/\mathcal{M}/\mathcal{C}$ subscripts in what follows and describe the generic MCA procedure for a single role.

\paragraph{Instance-level reflection.}
For each instance, MCA compresses the role-specific trace into a lightweight, attributable reflection:
\begin{equation}
\label{eq:reflection}
r_{t}
= \mathrm{Ref}\!\left(x_t, \mathcal{T}_{t}\right).
\end{equation}
Here $\mathrm{Ref}$ is a deterministic extractor that produces a short structured record from $\mathcal{T}_{t}$, including (i) iteration count $K_t$, (ii) outcome-aware quality tags (e.g., \texttt{task\_quality}, \texttt{task\_outcome}, and role-specific \texttt{quality} levels), and (iii) role-wise episode summaries of answers/diagnoses/actions across iterations.
These reflections serve as the inputs to subsequent reflection distillation.

\paragraph{Batch-level micro-lesson.}
To facilitate experience accumulation across multiple instances, we partition the reasoning stream into batches $\mathcal{B}_b$ of size $m$.
Within each batch $\mathcal{B}_b$, reflections are abundant and vary in usefulness.
We therefore perform \emph{in-batch distillation} to select informative reflections and summarize them into fewer micro-lessons.
We use a simple heuristic that leverages the MRO outcomes without requiring gold labels:
\begin{equation}
\label{eq:filter}
\begin{split}
\mathcal{S}_b^{+}&=\{t\in\mathcal{B}_b: K_t=1 \land a_t^{0}=\textsc{Accept}\},\\
\mathcal{S}_b^{-}&=\{t\in\mathcal{B}_b: K_t=N \land a_t^{N-1}=\textsc{Restart}\}.
\end{split}
\end{equation}
Here $\mathcal{S}_b^{+}$ contains ``best'' instances that succeed immediately, while $\mathcal{S}_b^{-}$ contains ``worst'' instances that exhaust the inner-loop budget and still require restarting.
We then distill reflections from both sets into a concise batch-level micro-lesson:
\begin{equation}
\label{eq:micro}
\ell_{b}
= \mathrm{Distill}\!\left(\{r_{t}\}|{t\in \mathcal{S}_b^{+}\cup \mathcal{S}_b^{-}}\right).
\end{equation}
Including both successful and failed cases provides complementary signals: successes suggest reusable tactics, while failures highlight negative patterns to avoid.

\paragraph{Global-level meta-knowledge.}
Micro-lessons should inform a long-term meta-knowledge state, but naively accumulating all past lessons can cause unbounded growth and outdated rules.
Inspired by recent findings on bounded test-time memory and multi-timescale learning~\cite{behrouz2025titans,behrouz2025nested}, we adopt \emph{temporal consolidation}.
Let $\mathrm{Win}_w(\cdot)$ keep the most recent $w$ batches of micro-lessons.
We update global-level meta-knowledge as:
\begin{equation}
\label{eq:meta}
K_{b}
= \mathrm{Consol}\!\left(K_{b-1},\mathrm{Win}_w(\{\ell_{j}\}_{j=1}^{b})\right).
\end{equation}
This bounded update balances retention and forgetting: recent, repeatedly supported lessons are preserved, while stale or low-utility rules naturally decay.

\paragraph{Updating MRO policies.}
After updating the meta-knowledge at the end of batch $\mathcal{B}_{b}$, we apply it to guide inference in the next batch $\mathcal{B}_{b+1}$.
Concretely, for each instance $x_i\in\mathcal{B}_b$, the MRO uses an updated policy prompt that integrates the current global meta-knowledge together with instance-relevant micro-lessons, thereby conditioning instance-level meta-reasoning on accumulated experience from previous batches.
The retrieval step is:
\begin{equation}
\label{eq:retrieve}
\tilde{\mathcal{L}}_{t} =
\mathrm{Retrieve}\!\left(x_i,\,\mathcal{L}^{(w)}_{b}\,|\,k\right).
\end{equation}
where $\mathcal{L}^{(w)}_{b}$ is windowed batches of micro-lessons.
The instance-conditioned prompt compilation is:
\begin{equation}
\label{eq:prompt_update}
P_{t+1} \leftarrow
\mathrm{update}\!\left(P_t, K_{b}, \tilde{\mathcal{L}}_{t},x_i\right).
\end{equation}

The overall algorithm is summarized in the Appendix~\ref{sec:appendix_algo}. All the policy prompts in MRO and procedure prompts in MCA are summarized in the Appendix~\ref{sec:appendix_prompt}.

\section{Experimental Setup}
\label{sec:exp_setup}

\paragraph{Datasets.}
We evaluate MC$^2$ on a suite of mathematical and symbolic reasoning benchmarks.
Our main experiments use four datasets: \textbf{GSM8K}~\cite{cobbe2021training}, a grade-school math word-problem benchmark emphasizing multi-step arithmetic;
\textbf{MATH-500}~\cite{hendrycks2021measuring}, a challenging subset of the MATH dataset covering diverse competition-style problems;
\textbf{TheoremQA}~\cite{chen-etal-2023-theoremqa}, a theorem-centric QA benchmark that tests formal and conceptual mathematical reasoning; and
\textbf{Game-of-24}~\cite{yao2023tree}, a symbolic search task where models must construct valid arithmetic expressions to reach a target value.

\paragraph{Baselines.}
We compare our framework against both standard reasoning methods and representative meta-reasoning and memory-based approaches.
For \textbf{standard reasoning}, we include Chain-of-Thought (\textbf{CoT})~\cite{wei2022chain}, CoT with Self-Consistency (\textbf{CoT-SC})~\cite{wang2023selfconsistency}, and Tree-of-Thought (\textbf{ToT})~\cite{yao2023tree}.
For \textbf{meta-reasoning and memory-based} baselines, we include \textbf{Meta-Prompting}~\cite{suzgun2024meta}, \textbf{Meta-Reasoner}~\cite{sui2025metareasoner},
Test-time Prompt Intervention (\textbf{TTPI})~\cite{yang2025test}, and
\textbf{Buffer-of-Thought}~\cite{yang2024buffer}.
To ensure a fair comparison, we \emph{re-implement} all baselines under a unified evaluation protocol.
Full implementation and hyperparameter details are reported in Appendix~\ref{sec:appendix_baselines}.

\paragraph{Backbone Models.}
We evaluate all methods on four backbone models to test robustness across model families and access regimes.
Specifically, we consider two close-sourced LLMs, \textbf{GPT-4o-mini} and \textbf{Gemini-2.0-flash}, and two open-weight instruction-tuned models, \textbf{Llama-3-8B-Instruct} and \textbf{Qwen3-8B}.
We keep decoding settings consistent across methods whenever applicable.
Exact prompting templates and decoding hyperparameters are summarized in Appendix~\ref{sec:appendix_implementation}.
\paragraph{Evaluation Metrics.}
We use accuracy as the primary evaluation metric across all benchmarks, and additionally report the number of decoding tokens as a measure of efficiency.
To reduce randomness and improve statistical reliability, we run each method three times with different random seeds and report the mean performance along with 95\% confidence intervals.


\begin{table*}[t]
\centering

\providecommand{\acc}[2]{\mbox{#1~($\pm#2$)}}
\providecommand{\bestacc}[2]{\textbf{\mbox{#1~($\pm#2$)}}}
\providecommand{\secondacc}[2]{\mbox{\underline{#1}~($\pm#2$)}}

\providecommand{\tok}[2]{\mbox{#1~($\pm#2$)}}

\small
\setlength{\tabcolsep}{6pt}
\renewcommand{\arraystretch}{1.12}

\resizebox{2\columnwidth}{!}{
\begin{tabular}{>{\raggedright\arraybackslash}p{3.2cm} *{8}{>{\centering\arraybackslash}p{1.65cm}}}
\toprule\toprule
& \multicolumn{2}{c}{GSM8K}
& \multicolumn{2}{c}{MATH-500}
& \multicolumn{2}{c}{TheoremQA}
& \multicolumn{2}{c}{Game-of-24} \\
\cmidrule(lr){2-3}\cmidrule(lr){4-5}\cmidrule(lr){6-7}\cmidrule(lr){8-9}
& Accuracy & \#Tokens
& Accuracy & \#Tokens
& Accuracy & \#Tokens
& Accuracy & \#Tokens \\
\midrule

\multicolumn{9}{c}{\textit{GPT-4o-mini}} \\
\addlinespace[2pt]
CoT               & \acc{91.38}{0.60}          & \tok{358}{2}    & \acc{72.67}{2.01}          & \tok{625}{14}  & \acc{44.04}{1.70}          & \tok{716}{1}    & \acc{7.67}{1.43}            & \tok{553}{15}   \\
CoT-SC            & \secondacc{94.44}{2.26}    & \tok{1805}{18}  & \secondacc{78.93}{1.25}    & \tok{3200}{37} & \secondacc{46.92}{0.95}    & \tok{3643}{4}   & \acc{27.00}{0.00}           & \tok{2461}{28}  \\
ToT               & \acc{94.41}{0.44}          & \tok{1016}{78}  & \acc{77.60}{3.44}          & \tok{3473}{579}& \acc{45.33}{0.95}          & \tok{5774}{200} & \acc{33.33}{5.17}           & \tok{3801}{494} \\
Meta Reasoning    & \acc{93.96}{0.57}          & \tok{361}{7}    & \acc{76.40}{1.49}          & \tok{657}{16}  & \acc{43.62}{1.35}          & \tok{752}{8}    & \acc{20.33}{7.99}           & \tok{964}{151}  \\
Meta Prompt       & \acc{93.40}{1.23}          & \tok{1042}{25}  & \acc{75.73}{3.31}          & \tok{1727}{149}& \acc{44.62}{1.64}          & \tok{2265}{70}  & \secondacc{41.33}{6.25}     & \tok{3178}{474} \\
Buffer of Thought & \acc{93.78}{1.00}          & \tok{1373}{5}   & \acc{75.33}{1.03}          & \tok{1756}{19} & \acc{44.42}{3.17}          & \tok{1867}{15}  & \acc{19.33}{3.79}           & \tok{1539}{8}   \\
TTPI              & \acc{92.50}{1.47}          & \tok{874}{20}   & \acc{74.33}{3.53}          & \tok{1208}{72} & \acc{44.42}{2.35}          & \tok{1484}{67}  & \acc{28.00}{4.30}           & \tok{916}{183}  \\
MC$^2$            & \bestacc{96.92}{0.11}      & \tok{833}{21}   & \bestacc{85.13}{1.74}      & \tok{1631}{129}& \bestacc{55.96}{1.47}      & \tok{2159}{143} & \bestacc{88.67}{8.72}       & \tok{2479}{542} \\
\addlinespace[2pt]
\midrule
\addlinespace[2pt]

\multicolumn{9}{c}{\textit{Gemini-2.0-flash}} \\
\addlinespace[2pt]
CoT               & \acc{95.58}{0.72}          & \tok{218}{4}    & \acc{91.53}{0.29}          & \tok{602}{42}  & \acc{59.34}{0.89}          & \tok{813}{26}   & \acc{72.00}{7.45}           & \tok{599}{75}   \\
CoT-SC            & \secondacc{96.03}{0.48}    & \tok{1100}{5}   & \acc{94.20}{2.77}          & \tok{3018}{161}& \acc{60.12}{1.13}          & \tok{4025}{45}  & \acc{80.67}{7.59}           & \tok{3085}{55}  \\
ToT               & \acc{95.98}{0.37}          & \tok{558}{54}   & \secondacc{94.93}{2.35}    & \tok{1999}{162}& \secondacc{61.33}{1.56}    & \tok{4743}{201} & \acc{83.33}{1.43}           & \tok{5549}{887} \\
Meta Reasoning    & \acc{95.83}{0.94}          & \tok{219}{2}    & \acc{91.73}{1.15}          & \tok{638}{46}  & \acc{59.54}{4.35}          & \tok{860}{40}   & \acc{81.00}{4.30}           & \tok{1682}{547} \\
Meta Prompt       & \acc{95.43}{0.66}          & \tok{1111}{135} & \acc{91.07}{1.03}          & \tok{1465}{147}& \acc{57.50}{1.12}          & \tok{1832}{175} & \secondacc{88.00}{2.48}     & \tok{1592}{269} \\
Buffer of Thought & \acc{95.96}{0.39}          & \tok{1324}{4}   & \acc{92.27}{3.31}          & \tok{2192}{64} & \acc{59.17}{2.41}          & \tok{2532}{10}  & \acc{78.00}{7.45}           & \tok{2158}{117} \\
MC$^2$            & \bestacc{97.17}{0.57}       & \tok{573}{23}   & \bestacc{96.73}{0.57}      & \tok{1299}{111}& \bestacc{68.04}{2.92}      & \tok{1753}{22}  & \bestacc{94.33}{1.43}       & \tok{1219}{216} \\
\addlinespace[2pt]
\midrule
\addlinespace[2pt]

\multicolumn{9}{c}{\textit{Qwen3-8B}} \\
\addlinespace[2pt]
CoT               & \acc{93.45}{1.21}          & \tok{305}{2}    & \acc{84.40}{2.17}          & \tok{932}{37}  & \acc{54.33}{1.17}          & \tok{1022}{44}  & \acc{48.00}{4.30}           & \tok{2960}{492} \\
CoT-SC            & \acc{94.41}{0.29}          & \tok{1537}{8}   & \acc{87.93}{0.76}          & \tok{4199}{63} & \acc{55.25}{0.30}          & \tok{4882}{50}  & \secondacc{68.67}{1.43}     & \tok{7325}{385} \\
ToT               & \acc{94.09}{0.39}          & \tok{877}{52}   & \acc{87.53}{0.76}          & \tok{3232}{203}& \acc{57.29}{1.00}          & \tok{5686}{243} & \acc{61.67}{12.25}          & \tok{6562}{636} \\
Meta Reasoning    & \acc{94.06}{0.29}          & \tok{306}{6}    & \acc{85.27}{1.15}          & \tok{978}{89}  & \acc{49.92}{0.95}          & \tok{888}{36}   & \acc{47.33}{1.43}           & \tok{2818}{227} \\
Meta Prompt       & \secondacc{94.62}{0.37}    & \tok{522}{12}   & \secondacc{91.93}{1.52}    & \tok{1545}{22} & \secondacc{61.29}{0.19}    & \tok{1448}{30}  & \acc{48.00}{4.97}           & \tok{7393}{4340}\\
Buffer of Thought & \acc{93.68}{0.39}          & \tok{1316}{8}   & \acc{86.60}{1.31}          & \tok{1983}{71} & \acc{54.44}{0.68}          & \tok{2171}{22}  & \acc{45.67}{5.17}           & \tok{2880}{364} \\
TTPI              & \acc{89.74}{1.09}          & \tok{300}{3}    & \acc{83.73}{0.29}          & \tok{888}{140} & \acc{52.29}{2.63}          & \tok{907}{107}  & \acc{58.67}{12.25}          & \tok{6410}{1453}\\
MC$^2$            & \bestacc{96.66}{0.66}      & \tok{726}{32}   & \bestacc{93.07}{1.60}      & \tok{1582}{242}& \bestacc{64.88}{0.93}      & \tok{1924}{164} & \bestacc{80.67}{3.79}       & \tok{4445}{1838}\\
\addlinespace[2pt]
\midrule
\addlinespace[2pt]

\multicolumn{9}{c}{\textit{Llama-3.1-8B-Instruct}} \\
\addlinespace[2pt]
CoT               & \acc{85.39}{1.44}          & \tok{528}{126}  & \acc{56.27}{1.25}          & \tok{1942}{1012}& \acc{39.50}{1.24}         & \tok{1097}{312} & \acc{16.67}{7.59}           & \tok{14734}{2112}\\
CoT-SC            & \acc{87.29}{0.95}          & \tok{1936}{204} & \acc{61.47}{3.19}          & \tok{7652}{424} & \acc{43.08}{2.87}         & \tok{8010}{231} & \acc{24.00}{6.57}           & \tok{30741}{2086}\\
ToT               & \acc{87.67}{0.76}          & \tok{1478}{106} & \acc{60.47}{2.35}          & \tok{6450}{1036}& \acc{43.79}{0.89}         & \tok{8296}{684} & \acc{33.00}{4.97}           & \tok{15249}{706} \\
Meta Reasoning    & \acc{86.55}{1.92}          & \tok{447}{24}   & \acc{54.67}{2.91}          & \tok{2117}{526} & \acc{48.80}{1.56}         & \tok{2052}{111} & \acc{32.67}{1.43}           & \tok{12631}{998} \\
Meta Prompt       & \secondacc{90.44}{1.85}    & \tok{1217}{310} & \secondacc{65.47}{0.29}    & \tok{4426}{677} & \secondacc{53.79}{3.62}   & \tok{3709}{1054}& \secondacc{41.00}{8.96}     & \tok{22895}{5859}\\
Buffer of Thought & \acc{86.68}{0.61}          & \tok{1745}{367} & \acc{55.80}{3.58}          & \tok{5564}{472} & \acc{49.92}{1.40}         & \tok{2662}{93}  & \acc{28.67}{3.79}           & \tok{6146}{352}  \\
TTPI              & \acc{90.17}{0.72}          & \tok{1426}{278} & \acc{54.80}{16.36}         & \tok{4284}{1327}& \acc{36.29}{2.82}         & \tok{3757}{839} & \acc{33.00}{4.30}           & \tok{15106}{3139}\\
MC$^2$            & \bestacc{95.55}{1.87}      & \tok{1320}{585} & \bestacc{76.80}{8.36}      & \tok{4435}{1955}& \bestacc{61.46}{2.29}     & \tok{4899}{992} & \bestacc{45.33}{5.74}       & \tok{12808}{2935}\\

\bottomrule\bottomrule
\end{tabular}
}
\caption{Main results (with 95$\%$ confidence intervals). Best/second-best are highlighted by \textbf{bold}/\underline{underline} in Accuracy. Tokens are rounded to integers.}
\label{tab:main_results}
\end{table*}

\section{Experimental Results}
\label{sec:exp_results}

In this section, we present comprehensive experimental results and analyses to assess the effectiveness of MC$^2$.
We organize our study around the following research questions (\textbf{RQs}):
\begin{itemize}
    \item \textbf{RQ1}: How does our framework perform compared to standard reasoning and meta-reasoning baselines across datasets and backbone models?
    \item \textbf{RQ2}: Does the framework exhibit \emph{improved reasoning performance over time}?
    \item \textbf{RQ3}: How do different architectural components (MRO, MCA, and their variants) contribute to the overall performance?
    \item \textbf{RQ4}: What does the learned meta-knowledge capture, and how does it evolve with time?
\end{itemize}

\begin{figure*}[t]
    \centering
    \includegraphics[width=0.7\textwidth]{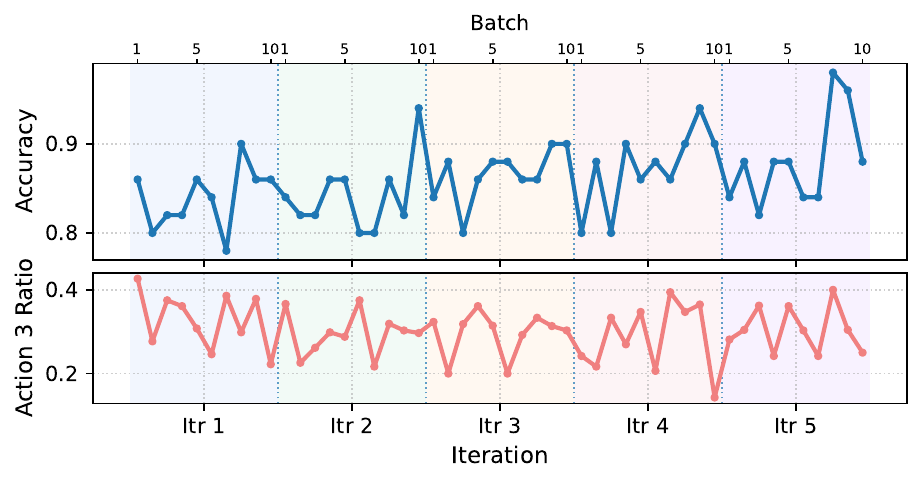}
    \caption{Top: Accuracy across batches and iterations. Bottom: Ratio of severe restart actions (Action 3) across batches and iterations.}
    \label{fig:time_evolvement}
\end{figure*}

\subsection{Main Results}
\label{sec:main_results}

To answer \textbf{RQ1}, we compare MC$^2$ with standard reasoning baselines (CoT, CoT-SC, ToT) and representative meta-reasoning/memory methods on GSM8K, MATH-500, TheoremQA, and Game-of-24 across four backbone models.
Table~\ref{tab:main_results} reports accuracy (with 95\% confidence intervals) and token usage.
We summarize four key findings.

\paragraph{Consistent improvements across datasets and backbones.}
MC$^2$ achieves the best accuracy across all backbone \& benchmark combinations, demonstrating that metacognitive consolidation generalizes well across tasks and models.
Representative gains include GPT-4o-mini on MATH-500 ($78.93\!\rightarrow\!85.13$) and TheoremQA ($46.92\!\rightarrow\!55.96$), as well as Llama-3.1-8B-Instruct on MATH-500 ($65.47\!\rightarrow\!76.80$).
These results indicate that consolidating meta-reasoning experience yields robust performance improvements.

\paragraph{Significant gains on control-intensive tasks.}
Improvements on GSM8K are relatively modest due to strong baseline performance, whereas MC$^2$ shows substantial advantages on benchmarks that require verification and backtracking.
The effect is most significant on Game-of-24, where MC$^2$ outperforms the strongest baseline on GPT-4o-mini ($41.33\!\rightarrow\!88.67$).
This suggests that MC$^2$ is particularly effective when episodic meta-reasoning repeatedly incurs similar correction costs.

\paragraph{Stronger benefits for weaker backbones.}
MC$^2$ provides larger relative improvements for weaker backbones while still benefiting stronger ones.
For example, although Gemini-2.0-flash already performs well on MATH-500, MC$^2$ still improves accuracy from $94.93\%$ to $96.73\%$.
This pattern indicates that structured meta-reasoning and cross-instance consolidation can compensate for limited intrinsic robustness while complementing strong base models.

\paragraph{Good accuracy-efficiency trade-off.}
While MC$^2$ incurs higher token usage than one-pass baselines, it is more efficient than compute-heavy test-time scaling methods.
On GPT-4o-mini MATH-500, MC$^2$ uses $1631$ tokens compared to $3200$ (CoT-SC) and $3473$ (ToT), while achieving higher accuracy.
Compared to memory-based approaches such as Buffer-of-Thought, MC$^2$ often attains better accuracy with comparable or lower token usage, striking a practical balance between performance and efficiency.

\subsection{Time Evolvement Results}
\label{sec:learning_over_time}

To answer \textbf{RQ2}, we evaluate how performance and efficiency evolve as MC$^2$ processes more data and accumulates metacognitive knowledge over time.
Specifically, we run MC$^2$ on the MATH-500 with GPT-4o-mini for multiple passes, and record accuracy and action statistics across both batches and iterations.
Due to the heterogeneous difficulty of instances across batches, batch-level performance exhibits noticeable fluctuations, even though experience is continuously accumulated.

Despite batch-level variability, iteration-level trends show clear and consistent improvement.
As shown in Figure~\ref{fig:time_evolvement}, overall accuracy increases steadily from about $84\%$ at iteration~1 to $88\%$ at iteration~5, while the proportion of severe restart actions (Action~3) decreases from $33.1\%$ to $30.8\%$.
These results indicate that MC$^2$ becomes both more accurate and more stable as metacognitive knowledge accumulates.
Importantly, this improvement reflects a form of \emph{batch-level scaling}: instead of allocating more compute within a single instance, MC$^2$ amortizes computation across instances within a batch, enabling performance gains even without access to ground-truth supervision.

\begin{table}[t]
\centering
\small
\resizebox{0.85\columnwidth}{!}{
\begin{tabular}{lcc}
\toprule
\textbf{Method} & \textbf{Math-500} & \textbf{TheoremQA} \\
\midrule
\multicolumn{3}{c}{\textbf{GPT-4o-mini}} \\
\midrule
Only Reasoner             & 72.67 & 44.04 \\
Reasoner + Updates       & 80.47 & 50.46 \\
R+ M + C                 & 79.20 & 47.38 \\
R+ M + C + Micro-update  & 82.80 & 51.75 \\
Full Method               & \textbf{85.13} & \textbf{55.96} \\
\midrule
\multicolumn{3}{c}{\textbf{Qwen-3-8B}} \\
\midrule
Only Reasoner             & 84.40 & 54.33 \\
Reasoner + Updates       & 90.80 & 59.79 \\
R+ M + C                 & 91.33 & 59.62 \\
R+ M + C + Micro-update  & 92.60 & 62.21 \\
Full Method               & \textbf{93.07} & \textbf{64.88} \\
\bottomrule
    \end{tabular}}
\caption{Ablation study on Math-500 and TheoremQA under different backbone models.}
\label{tab:ablation}
\end{table}

\subsection{Ablation Studies}
\label{sec:ablation}

To answer \textbf{RQ3}, we conduct ablation studies to disentangle the effects of instance-level meta-reasoning (MRO) and cross-instance consolidation (MCA).
We evaluate five variants: \textit{Only Reasoner}, \textit{Reasoner + Updates}, \textit{R+M+C}, \textit{R+M+C+Micro-update}, and the \textit{Full Method}, on Math-500 and TheoremQA with two backbone models (Table~\ref{tab:ablation}).

Overall, performance improves consistently as components are added.
Compared with \textit{Only Reasoner}, \textit{Reasoner + Updates} already delivers clear gains on both backbones and both datasets, showing that prompt-level meta-knowledge updates alone provide useful guidance.
Enabling the full MRO loop (\textit{R+M+C}) yields competitive gains over the same baseline, showing that structured monitoring and control directly enhance inference.
Adding a simple memory mechanism (\textit{Micro-update}) further improves accuracy, while the \textit{Full Method} performs best, demonstrating the effectiveness of MCA's hierarchical consolidation.

\subsection{Case Study}
\label{sec:case_study}
To answer \textbf{RQ4}, we present a representative case on MATH-500 with GPT-4o-mini as backbone in Figure~\ref{fig:case_study}.
In this example, the Reasoner produces a wrong answer when prompted without meta-knowledge.
After incorporating the learned meta-knowledge and micro-lessons into the prompt, the Reasoner directly arrives at the correct solution without requiring additional intervention.
This case demonstrates the effectiveness of meta-knowledge-guided prompt rewriting in improving reasoning accuracy.
Additional examples are provided in Appendix~\ref{app:guidance-relevance}.

\begin{strip}
\centering
\includegraphics[width=0.85\textwidth]{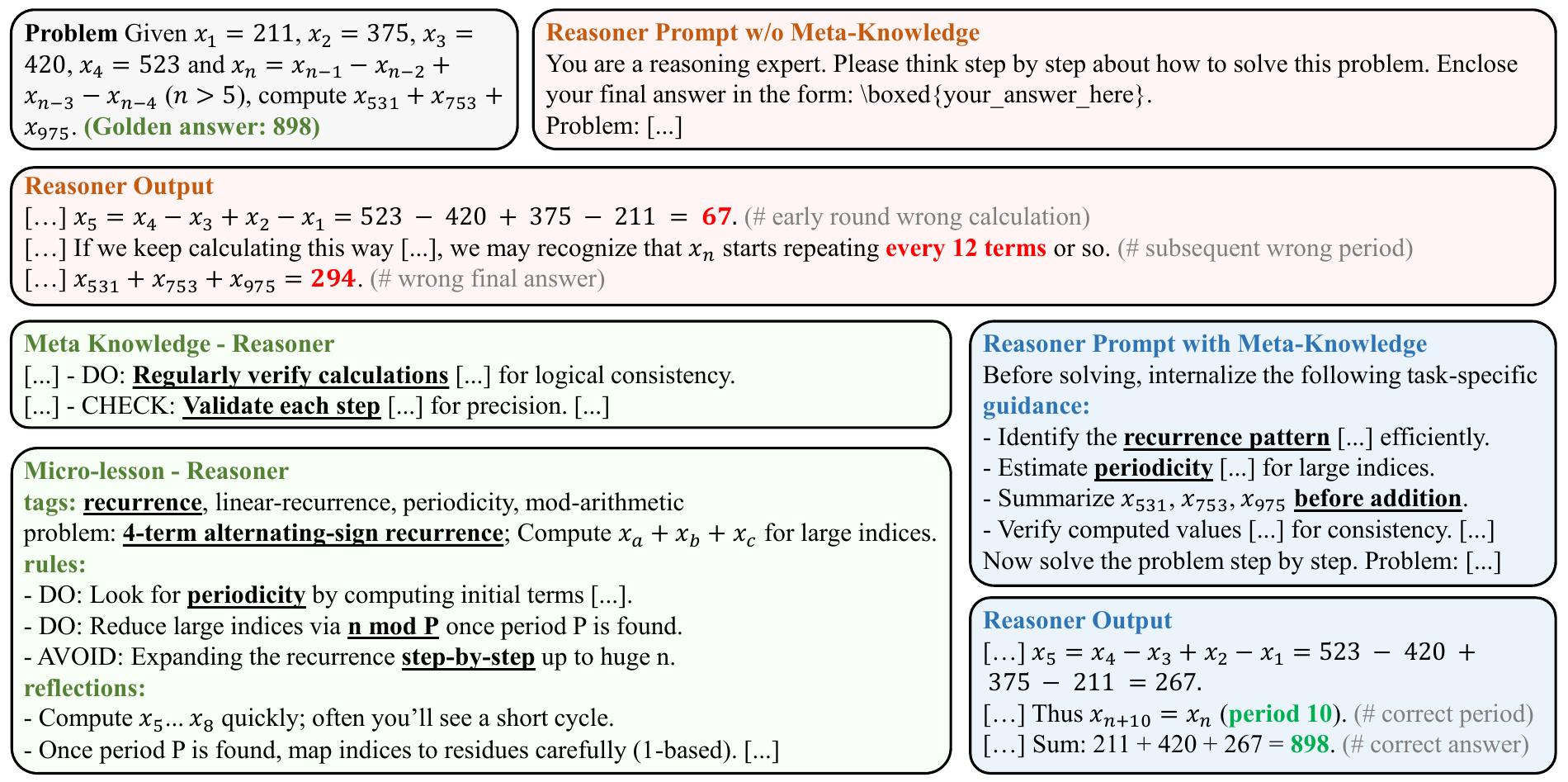}
\begingroup
\captionsetup{hypcap=false}
\captionof{figure}{Case study. \textcolor{red}{Red} text highlights erroneous steps or final answers, while \textcolor{green}{green} text indicates corrected reasoning and the correct answer. \textcolor{gray}{Gray} text denotes comments or explanations. For clarity, less relevant content is omitted and replaced with \texttt{[...]}.}
\label{fig:case_study}
\endgroup
\end{strip}

\section{Related Work}

\paragraph{LLM Meta-Reasoning}
Meta-reasoning refers to reasoning about how to reason, providing an additional axis for improving LLM reasoning.
Most work can be grouped into two paradigms: \textit{pre-task meta-planning} and \textit{in-task meta-control}.
Meta-Prompting~\cite{suzgun2024meta} and related scaffolding methods~\cite{gao2024meta} structure subsequent reasoning, while ReMA~\cite{wan2025rema} learns to generate meta-level guidance via multi-agent learning.
Other work maintains and evolves a pool of meta-thoughts for test-time scaling and selection~\cite{liu2025metascale}.
In contrast, in-task meta-control injects meta-level interventions during reasoning.
Test-time Prompt Intervention~\cite{yang2025test} modifies the reasoning process based on online signals.
RLVMR~\cite{zhang2025rlvmr} encourages verification and re-planning via verifiable rewards.
Hierarchical frameworks such as Thinker~\cite{xu2025thinker} coordinate multi-turn search, while Meta-Reasoner~\cite{sui2025metareasoner} provides dynamic inference-time guidance.
Recent position paper has also highlighted the need for principled meta-reasoning from a Bayesian perspective~\cite{yan2025position}.
While prior methods focus on \textit{episodic} meta-reasoning within individual instances, our work targets \textit{Metacognitive Consolidation}, accumulating reusable meta-reasoning skills across instances as persistent meta-knowledge.

\paragraph{Memory for Reasoning.}
Another related direction improves reasoning by equipping models (or agents) with memory that reuses past reasoning artifacts.
For example, ReasoningBank~\cite{ouyang2025reasoningbank} and Reflexion~\cite{shinn2023reflexion} store experience from previous rollouts (e.g., trajectories and reflections) to guide future behavior.
Buffer of Thoughts~\cite{yang2024buffer} maintains a retrievable buffer of distilled thought templates, and ReasonFlux~\cite{zou2025reasonfluxprm} builds hierarchical libraries of reusable reasoning patterns to reduce redundant exploration.
These approaches primarily store \textit{task-level} reasoning content (solutions, templates, or heuristics for solving), whereas we consolidate \textit{meta-level} experience, how to monitor, critique, and control reasoning, into procedural meta-knowledge that progressively improves future meta-reasoning.

\section{Conclusion}
\label{sec:conclusion}
In this paper, we propose Metacognitive Consolidation, a learning framework that enables models to transform metacognitive experience from past reasoning episodes into reusable procedural knowledge for future meta-reasoning.
Our framework structures instance-level reasoning into explicit roles and consolidates the resulting meta-level traces across instances via hierarchical updates.
Experiments show consistent gains across benchmarks and models, with performance improving as experience accumulates.
Future work will explore extensions to multi-agent settings and mechanisms for internalizing consolidated meta-knowledge more deeply into model parameters.

\section*{Limitations}

We identify three key limitations of our framework.

First, our current formulation primarily consolidates \emph{task-specific} metacognitive knowledge by repeatedly processing instances drawn from the same benchmark or task distribution.
In contrast, human metacognition naturally transfers across tasks and domains.
An important direction for future work is to study \emph{task-agnostic} or cross-task metacognitive consolidation, where reusable meta-skills learned in one setting can generalize to and accelerate reasoning in new tasks.

Second, metacognitive knowledge in our framework is represented explicitly at test time, via externalized meta-knowledge and prompt-based policy updates, rather than being internalized into model parameters.
We adopt this design as a practical starting point because it keeps the learned content inspectable and avoids expensive fine-tuning.
Since the accumulated knowledge mainly takes the form of reusable reasoning habits, verification routines, and control policies, parameter-level or hybrid internalization remains a natural next step that we leave to future work.

Third, enabling metacognitive consolidation incurs additional inference overhead, as the framework introduces multiple reasoning roles and periodic consolidation steps.
Although our experiments demonstrate favorable performance--efficiency trade-offs relative to existing meta-reasoning methods, further optimization is needed.
In particular, exploring more efficient reasoning representations, such as implicit chains of thought, may help reduce computational cost while preserving the benefits of metacognitive consolidation.

\section*{Acknowledgment}

This work was supported in part by the UK Engineering and Physical Sciences Research Council through a Turing AI Fellowship (grant no. EP/V020579/1, EP/V020579/2), the Prosperity Partnership scheme (grant no. UKRI566), and the Taishan Scholars Program (No. TSQN202507242).

\bibliography{custom}

\appendix
\clearpage
\newpage

\section*{Appendix}
\section{Overall Algorithm}
\label{sec:appendix_algo}

Algorithm~\ref{alg:mc2} summarizes MC$^2$.
At a high level, we (i) retrieve top-$k$ micro-lessons and compile instance-conditioned role prompts before each inference (Eqs.~\eqref{eq:retrieve}--\eqref{eq:prompt_update}),
(ii) run MRO to produce per-instance traces $\mathcal{T}_t$ via iterative monitoring and control (Eqs.~\eqref{eq:reasoner}--\eqref{eq:trace}),
and (iii) consolidate traces across instances into multi-frequency memory (Eqs.~\eqref{eq:reflection}--\eqref{eq:meta}).

\begin{algorithm}[h]
\caption{Metacognitive Consolidation (MC$^2$)}
\label{alg:mc2}
\begin{algorithmic}[1]
\Require Dataset/stream $\mathcal{D}$, batch size $m$, max inner iters $N$, window size $w$, retrieval top-$k$
\State Initialize base role prompts $P_R,P_M,P_C$ and meta-knowledge states $\{K_{0,i}\}_{i\in\{R,M,C\}}$
\For{batch $b=1,2,\dots$}
    \For{each instance $x_t\in\mathcal{B}_b$}
        \For{role $i\in\{R,M,C\}$}
            \If{$b=1$}
                \State $\tilde{P}_{t,i}\leftarrow P_i$ \Comment{cold start: no retrieval / meta-knowledge}
            \Else
                \State $\mathcal{L}^{(w)}_{b-1,i}\leftarrow \mathsf{Win}_w(\{\ell_{j,i}\}_{j=1}^{b-1})$
                \State $\tilde{\mathcal{L}}_{t,i}\leftarrow \mathsf{Retrieve}_i(x_t;\mathcal{L}^{(w)}_{b-1,i},k)$ \Comment{top-$k$ micro-lessons}
                \State $\tilde{P}_{t,i} \leftarrow \mathsf{UpdPrompt}_i(P_i, K_{b-1,i}, \tilde{\mathcal{L}}_{t,i}, x_t)$
                \Comment{compile instance-conditioned role prompt; Eq.~\eqref{eq:prompt_update}}
            \EndIf
        \EndFor
        \State Run MRO for at most $N$ iterations to obtain $(\hat{y}_t,\mathcal{T}_t)$ using Eqs.~\eqref{eq:reasoner}--\eqref{eq:trace}
        \State \hspace{1.2em}with compiled prompts $\tilde{P}_t=(\tilde{P}_{t,R},\tilde{P}_{t,M},\tilde{P}_{t,C})$
        \For{role $i\in\{R,M,C\}$}
            \State Compute instance reflection $r_{t,i}$ via Eq.~\eqref{eq:reflection}
        \EndFor
    \EndFor
    \For{role $i\in\{R,M,C\}$}
        \State Select $\mathcal{S}_b^{+},\mathcal{S}_b^{-}$ via Eq.~\eqref{eq:filter}
        \State Distill micro-lesson $\ell_{b,i}$ via Eq.~\eqref{eq:micro}
        \State Update meta-knowledge $K_{b,i}$ via Eq.~\eqref{eq:meta}
    \EndFor
\EndFor
\end{algorithmic}
\end{algorithm}

\section{Baselines}
\label{sec:appendix_baselines}

This section describes our baseline re-implementations to minimize confounds between \emph{method} differences and \emph{implementation} differences.

We compare our framework against both standard reasoning methods and representative meta-reasoning and memory-based approaches.
For \textbf{standard reasoning}, we include Chain-of-Thought (\textbf{CoT})~\cite{wei2022chain}, which elicits step-by-step reasoning;
CoT with Self-Consistency (\textbf{CoT-SC})~\cite{wang2023selfconsistency}, which improves reliability via sampling and majority voting; and
Tree-of-Thought (\textbf{ToT})~\cite{yao2023tree}, which performs structured search over intermediate thoughts.
For \textbf{meta-reasoning} baselines, we include \textbf{Meta-Prompting}~\cite{suzgun2024meta}, which provides task-agnostic scaffolding to guide reasoning;
\textbf{Meta-Reasoner}~\cite{sui2025metareasoner}, which dynamically adjusts inference-time reasoning strategies based on the current reasoning state with a contextual multi-armed bandits;
Test-time Prompt Intervention (\textbf{TTPI})~\cite{yang2025test}, which injects interventions during inference to steer the reasoning trajectory; and
\textbf{Buffer-of-Thought}~\cite{yang2024buffer}, which maintains and retrieves reusable thought templates to guide new instances.
To ensure a fair comparison, we \emph{re-implement} all baselines under a unified evaluation protocol.

Since some prior works do not release full code/prompts or rely on inaccessible closed-source evaluation settings, we follow a best-effort reproduction strategy: we adhere to each method's core inference procedure (e.g., sampling, search, intervention, retrieval) and implement end-to-end inference under a unified input--output protocol.

\paragraph{Unified I/O protocol and answer parsing.}
All baselines share the same answer extraction and normalization.
Concretely, we only parse predictions from an explicit final-answer field (e.g., \texttt{Final Answer: <answer>}).
If the output is invalid or unparsable, we apply the method's prescribed fallback (e.g., retry/selection rules), and otherwise mark the attempt as failure under the shared parsing rule.
Unless stated otherwise, we keep the same backbone model, decoding settings, and stopping criteria across baselines to reduce confounds.

\vspace{0.5em}
\subsection{Standard reasoning baselines}

\paragraph{CoT (Chain-of-Thought).}
We implement the standard ``step-by-step reasoning $\rightarrow$ final answer'' pipeline.
The prompt instructs the model to produce intermediate reasoning and then output a single, parseable answer after a fixed marker (the final-answer field).
Evaluation parses only the final-answer field and applies basic normalization to match dataset annotations.

\paragraph{CoT-SC (Self-Consistency).}
Using the same CoT prompt, we draw $5$ independent samples per input and aggregate the parsed final answers.
We first normalize equivalent answers, then perform majority voting.
If some samples are unparsable, we deterministically select the candidate that is (i) parseable and (ii) has the highest support among parseable candidates (ties broken deterministically).

\paragraph{ToT (Tree-of-Thought).}
We reproduce ToT with a breadth-first tree search that iterates
\emph{(generate candidate thoughts $\rightarrow$ self-evaluate/rank $\rightarrow$ keep top-$k$ to expand)}
until reaching a maximum depth or a termination condition.
We use: $5$ candidates per node, top-$2$ expansion, and maximum depth $2$.
We replace task-specific evaluators with a generic self-evaluation prompt using the same backbone model, and record the generation/evaluation prompts and stopping rules.

\vspace{0.5em}
\subsection{Meta-reasoning and memory-augmented baselines}

\paragraph{Meta-Prompting.}
We reproduce the multi-role collaboration paradigm of meta-prompting: the same backbone LLM plays a Meta Model (controller) and multiple Expert roles, distinguished only by role-specific prompt templates.
Inference maintains a growing message history.
At each round, the Meta Model generates the next action from the full history; if it issues an expert instruction, we extract it and call the corresponding expert in an isolated format:
\texttt{Expert X: '''...'''}.
We then append the expert response back to history and continue.
If the Meta Model outputs the final-answer marker, we parse and return it; otherwise we follow the original procedure by appending an error message and iterating.
Expert calls follow the ``fresh eyes'' setting: experts only see the triple-quoted instruction, not the full history.
Unlike the original work, we do not implement the Python Expert / interpreter, as our benchmarks do not involve code; we therefore remove code-execution components and retain only the single-expert-per-round, task decomposition, and expert verification rules.

\paragraph{Meta-Reasoning.}
We adopt the seven reasoning strategies provided in the Meta-Reasoning work and use the same backbone model to select which strategy to apply at test time.
When the selected strategy is ToT or CoT-SC, we use the same hyperparameters as specified above.

\paragraph{TTPI (Test-time Prompt Intervention).}
We reproduce TTPI's dynamic intervention procedure by segmenting generation into step-level stages.
At the end of each stage, we construct multiple candidate continuations corresponding to different intervention trigger sets (e.g., favoring progression, summarization, verification, or conclusion) and select one branch as the next step, repeating until a final answer is produced.
TTPI requires per-token log probabilities to compute perplexity (PPL), and the full version also uses internal-layer differences to compute the Reasoning Depth Score (RDS) for branch selection.
Due to API constraints, we do not evaluate TTPI on Gemini-2.0-flash (no per-token log probabilities).
On GPT-4o-mini, per-token log probabilities are available but internal-layer access is not; thus, in the \emph{Which} module we select branches using PPL only, while keeping the rest of the intervention procedure and stopping rules unchanged.

\paragraph{Buffer-of-Thought (BoT).}
We reproduce BoT's \emph{distill--retrieve--instantiate--update} pipeline.
We first apply a \emph{Problem Distiller} to convert each problem into a structured distilled representation (e.g., key facts, goals, constraints) for retrieval and reasoning.
We then retrieve the most relevant thought template from the Meta-buffer by embedding similarity between the distilled representation and template descriptions; if similarity is below a threshold, we treat it as a new task and fall back to a generic coarse-grained template.
Next, an \emph{Instantiation} prompt combines the retrieved template with the distilled representation to guide structured reasoning and produce the final answer.
Finally, a \emph{Buffer-manager} extracts a reusable thought template from the solved trajectory and adds it to the Meta-buffer only if it is sufficiently novel under the same similarity criterion, enabling continual accumulation and reuse.

\begin{table}[H]
\centering
\small
\begin{tabularx}{\linewidth}{l X}
\hline
\textbf{Baseline} & \textbf{Key settings in our reproduction} \\
\hline
CoT & Single sample; parse only \texttt{Final Answer:} field \\
CoT-SC & 5 samples; normalize + majority vote; deterministic fallback \\
ToT & BFS; 5 candidates/node; top-2 expand; max depth 2; self-eval prompt \\
Meta-Prompting & Meta controller + experts via role prompts; iterative multi-round; experts see only isolated triple-quoted instructions (``fresh eyes''); stop on final-answer marker; no code-execution expert \\
Meta-Reasoning & Strategy selection; when strategy is ToT/CoT-SC use the same settings as above \\
TTPI & Step-stage interventions; branch selection via PPL when available (no RDS without internal-layer access) \\
BoT & Distill--retrieve--instantiate--update; similarity-threshold gating for template retrieval and buffer updates \\
\hline
\end{tabularx}
\vspace{-0.5em}
\caption{Hyperparameter and procedure summary for baseline reproductions.}
\label{tab:baseline_hparams}
\end{table}

\section{Implementation}
\label{sec:appendix_implementation}
\subsection{Heuristic Construction of Instance-level Reflections}
\label{sec:appendix_instance_reflection}

For each instance, we construct an \emph{instance-level reflection} by aggregating signals from its interaction trace across the \textsc{Reasoner}--\textsc{Monitor}--\textsc{Controller} loop. The reflection is a compact, structured summary of (i) the task outcome, (ii) how many inner-loop iterations were required, and (iii) coarse-grained role-level quality ratings. It is derived only from trace-visible fields (plus the correctness label) and does not require additional model calls. We use the correctness flag only as a binary outcome tag for reflection labeling; the gold answer text $y_t^\star$ is never provided to any LLM call during reflection, distillation, retrieval, or consolidation.

\paragraph{Inputs extracted from the trace.}
Let the trace contain $T$ iteration records $\mathcal{H}=\{h_1,\dots,h_T\}$. From each iteration $h_t$, we extract:
(i) the monitor's \emph{error-found flag} $e_t$,
(ii) the controller's \emph{action code} $a_t$,
and lightweight per-iteration facts for constructing each agent's \emph{behavior trajectory} (defined below).
We also use instance-level metadata: terminal status $s$ (e.g., accepted/corrected vs.\ max-iteration) and correctness flag $y\in\{0,1\}$.
We denote the monitor's final-iteration flag by $e_T$.

We summarize two trace-level counts:
\begin{equation}
\label{eq:trace-counts}
m_{\texttt{YES}}=\sum_{t=1}^{T}\mathbb{I}[e_t=\texttt{YES}],\qquad
c_{3}=\sum_{t=1}^{T}\mathbb{I}[a_t=3],
\end{equation}
where $m_{\texttt{YES}}$ counts how often the monitor flags an error, and $c_3$ counts how often the controller selects the restart action.

\paragraph{Structured reflection fields.}
Each instance-level reflection contains:
\begin{itemize}
  \item \textbf{Task outcome:} a binary label (\texttt{success} or \texttt{failure}) derived from the correctness flag.
  \item \textbf{Task quality:} a three-level label (\texttt{A}/\texttt{B}/\texttt{C}) describing whether the system solved the instance cleanly in one iteration, required multiple iterations, or failed to converge.
  \item \textbf{Reasoner quality:} a three-level rating described as \emph{good}, \emph{ok}, or \emph{poor} (stored as \texttt{R\_good}, \texttt{R\_ok}, \texttt{R\_poor} in logs), derived from termination type, iteration count, and the prevalence of monitor-flagged errors.
  \item \textbf{Monitor quality:} a three-level rating described as \emph{good}, \emph{ok}, or \emph{poor} (stored as \texttt{M\_good}, \texttt{M\_ok}, \texttt{M\_poor}), based on whether the monitor's final judgment is aligned with successful termination (e.g., the final iteration should not still flag an error if the run terminates as accepted/corrected).
  \item \textbf{Controller quality:} a three-level rating described as \emph{good}, \emph{ok}, or \emph{poor} (stored as \texttt{C\_good}, \texttt{C\_ok}, \texttt{C\_poor}), based on whether the controller converges efficiently and whether it over-uses restart actions in non-convergent runs.
  \item \textbf{Role behavior trajectories:} three compact per-role trajectories extracted from the trace, each recording the key observable actions taken at every iteration (in chronological order). Concretely:
  \begin{itemize}
    \item \textbf{Reasoner trajectory:} the sequence of extracted final answers $\{\hat{z}_t\}_{t=1}^{T}$ (and optionally a short excerpt pointer to the full response text for logging).
    \item \textbf{Monitor trajectory:} the sequence $\{(e_t,\mathrm{step}_t,\mathrm{desc}_t)\}_{t=1}^{T}$, where $e_t$ is the error-found flag and $\mathrm{step}_t,\mathrm{desc}_t$ are the reported error location and short description (when available).
    \item \textbf{Controller trajectory:} the sequence $\{(a_t,\mathrm{just}_t)\}_{t=1}^{T}$, where $a_t$ is the action code and $\mathrm{just}_t$ is the controller's justification string (or a brief extract).
  \end{itemize}
\end{itemize}

\paragraph{Heuristic rules.}
\begin{itemize}
  \item \textbf{Task outcome.} We set task outcome to \texttt{success} if $y=1$; otherwise \texttt{failure}.
  \item \textbf{Task quality.} We set task quality to \texttt{A} if $s$ indicates accepted/corrected and $T=1$, to \texttt{B} if $s$ indicates accepted/corrected and $T>1$, and to \texttt{C} otherwise (including max-iteration termination).
  \item \textbf{Reasoner quality.} We rate the reasoner as \emph{good} if the run is accepted/corrected within two iterations; as \emph{poor} if the run hits the maximum iteration budget and more than half of iterations are flagged as erroneous by the monitor (i.e., $m_{\texttt{YES}} > T/2$); otherwise as \emph{ok}.
  \item \textbf{Monitor quality.} For accepted/corrected runs, we rate the monitor as \emph{good} if the final iteration does not flag an error ($e_T=\texttt{NO}$), and as \emph{poor} if it still flags an error ($e_T=\texttt{YES}$); otherwise we assign \emph{ok}. For non-accepted runs, we assign \emph{ok}.
  \item \textbf{Controller quality.} We rate the controller as \emph{good} if the run is accepted/corrected within three iterations; as \emph{poor} if the run hits the maximum iteration budget and more than half of iterations choose the restart action (i.e., $c_3 > T/2$); otherwise as \emph{ok}.
\end{itemize}

\paragraph{Optional episode summaries (logging only).}
For debugging and analysis, we also produce three human-readable reflections (one per role), each including the auto-wrapping question brief and enumerating the corresponding role behavior trajectory over iterations (e.g., the reasoner's final answer per iteration; the monitor's error flags/steps/descriptions; the controller's actions and justifications). These texts are not required by the reflection schema itself and do not affect downstream construction.

\subsection{Does first-try success align with correctness?}
\label{sec:appendix_first_try_alignment}

Using the task-quality tags defined above, we examine whether emphasizing first-try success risks optimizing for faster termination rather than correctness.
We analyze pooled results from three runs each of GPT-4o-mini and Gemini-2.0-flash on MATH-500 (six runs in total).
We find that first-try success is strongly aligned with correctness: Grade A instances achieve $95.95\%$ accuracy, whereas Grade C instances achieve only $38.82\%$, with a statistically significant positive correlation between better task quality and correctness (Spearman's $\rho=0.44$, $p<0.001$).

\begin{table}[t]
\centering
\small
\resizebox{0.98\columnwidth}{!}{
\begin{tabular}{ccccc}
\toprule
\textbf{Level} & \textbf{Samples} & \textbf{Overall Acc} & \textbf{Grade A Rate} & \textbf{Grade A Acc} \\
\midrule
1 & 258 & 97.29\% & 98.06\% & \textasciitilde{}98\% \\
2 & 540 & 97.59\% & 94.44\% & \textasciitilde{}98\% \\
3 & 630 & 97.14\% & 91.43\% & \textasciitilde{}97\% \\
4 & 768 & 91.67\% & 83.07\% & \textasciitilde{}95\% \\
5 & 804 & 78.86\% & 73.63\% & \textasciitilde{}94\% \\
\bottomrule
\end{tabular}}
\caption{Difficulty-level breakdown of first-try correctness on MATH-500, pooled over GPT-4o-mini and Gemini-2.0-flash across six runs.}
\label{tab:appendix_first_try_alignment}
\end{table}

Table~\ref{tab:appendix_first_try_alignment} further shows that Grade A is not restricted to easy questions.
Even on Level 5, the hardest subset of MATH-500, $73.63\%$ of instances are still Grade A, and those Grade A cases remain highly accurate at about \textasciitilde{}94\%.
This indicates that MC$^2$ is not simply filtering out easy problems faster; it is increasingly solving difficult problems correctly on the first try.
Moreover, Grade C cases are not discarded: in the distillation filter of Eq.~\eqref{eq:filter}, failed cases are explicitly retained through $\mathcal{S}_b^{-}$ as negative signals, allowing the system to learn not only from successful first-try behaviors but also from failure patterns that should be avoided.

\subsection{MRO role strategy Update}
\label{sec:appendix_prompt_update_impl}

The prompt update operator in Eq.~\eqref{eq:prompt_update} is implemented via an \emph{LLM-based prompt composer}.
Concretely, for each role (\textsc{Reasoner}/\textsc{Monitor}/\textsc{Controller}), we maintain a role-specific policy prompt at time $t$ and update it by composing:
(i) the previous policy prompt $P_t$,
(ii) the current global meta-knowledge state $K_b$,
(iii) the retrieved instance-relevant micro-lessons $\widetilde{\mathcal{L}}_{t}$ from Eq.~\eqref{eq:retrieve_impl},
and (iv) the current instance text $x_i$.
The composer itself is realized by a dedicated prompt template (see Appendix~\ref{sec:appendix_prompt}) and is executed by calling the backbone model $\mathrm{LLM}_\theta$.

\paragraph{Template and output format.}
We use a fixed \emph{prompt-composer template} defined in Appendix~\ref{sec:appendix_prompt} that instructs the model to produce updated role prompts in a pre-specified structured schema (e.g., with explicit role fields such as \texttt{P\_R}, \texttt{P\_M}, \texttt{P\_C}).
This schema is designed so that the updated prompts can be parsed deterministically and directly fed into the next MRO call.

\paragraph{Validity checks, regeneration, and fallback.}
After generation, we apply lightweight checks to ensure the composer output is usable: (i) all required role fields are present; (ii) the output is parseable under the expected schema; and (iii) the resulting prompt length does not exceed a pre-set context budget for the next inference call.
If any check fails, we re-run the prompt composer to regenerate the updated prompts .
If regeneration still fails, we fall back to using the previous prompt $P_t$ for that role (i.e., no update is applied for that step).

\paragraph{Trace logging.}
When a prompt update is attempted, we log the update metadata (whether an update was used, whether it succeeded, which model performed the update, and prompts before/after update) under the \textbf{Updated role policy} field in each role's trace record (Appendix~\ref{sec:appendix_trace_schema}).

\subsection{In-batch Distillation}
\label{sec:appendix_inbatch_distill}

The distillation operator $\mathrm{Distill}(\cdot)$ in Eq.~\eqref{eq:micro} is implemented as an \emph{LLM-based lesson distiller}.
For each role (\textsc{Reasoner}/\textsc{Monitor}/\textsc{Controller}), we run distillation \emph{independently} to obtain role-specific micro-lessons.
Given a batch $\mathcal{B}_b$, the distiller takes as input a set of instance-level reflections selected by the heuristic filter in Eq.~\eqref{eq:filter}, and outputs a batch-level micro-lesson $\ell_{b}$ that summarizes \emph{reusable tactics} (from $\mathcal{S}_b^{+}$) and \emph{failure patterns to avoid} (from $\mathcal{S}_b^{-}$).

\paragraph{Inputs.}
For batch $\mathcal{B}_b$, we form the distillation input set
$\{r_t\}_{t\in \mathcal{S}_b^{+}\cup \mathcal{S}_b^{-}}$,
where each reflection $r_t$ is constructed deterministically (Appendix~\ref{sec:appendix_instance_reflection}) and contains only trace-derived fields plus a binary outcome tag.
Importantly, we never provide the gold answer text $y_t^\star$ to the distiller; the correctness flag is used only as a coarse label (\texttt{success}/\texttt{failure}) inside reflections.

\paragraph{Prompt template and output format.}
We implement the distiller using a fixed prompt template (Appendix~\ref{sec:appendix_prompt}) executed by the backbone model $\mathrm{LLM}_\theta$.
The template specifies the target role and instructs the model to (i) identify recurring patterns across reflections, (ii) extract actionable rules/checks for the role, and (iii) summarize both positive (\emph{do}) and negative (\emph{avoid}) lessons grounded in the input reflections.
The distiller outputs $\ell_b$ in a pre-specified structured schema, as defined in the prompt.

\subsection{Temporal Consolidation (T-consolidation)}
\label{sec:appendix_tconsolidation}

The consolidation operator $\mathrm{Consol}(\cdot)$ in Eq.~\eqref{eq:meta} is implemented via an \emph{LLM-based meta-knowledge consolidator}.
The goal is to maintain an evolving global meta-knowledge state $K_b$ per role, while enabling natural forgetting through the sliding window $\mathrm{Win}_w(\cdot)$.

\paragraph{Inputs.}
At the end of batch $\mathcal{B}_b$, the consolidator receives:
(i) the previous global meta-knowledge $K_{b-1}$ for the target role, and
(ii) the windowed set of recent micro-lessons $\mathcal{L}_b^{(w)} \triangleq \mathrm{Win}_w(\{\ell_j\}_{j=1}^{b})$.
Only micro-lessons from the most recent $w$ batches are provided, so outdated rules are implicitly deprioritized without requiring explicit timestamps.

\paragraph{Prompt template and output format.}
We implement the consolidator using a fixed prompt template (Appendix~\ref{sec:appendix_prompt}) executed by $\mathrm{LLM}_\theta$.
The template instructs the model to consolidate $\mathcal{L}_b^{(w)}$ into an updated global state $K_b$ by merging redundant lessons, resolving minor inconsistencies, and rewriting the remaining rules into a concise, role-executable form.
The consolidator outputs $K_b$ in a pre-specified structured schema, as defined in the prompt, which can be directly injected into subsequent role-policy updates and guided inference.

\subsection{Trace Schemas}
\label{sec:appendix_trace_schema}

For each input instance, we record an ordered \emph{interaction trace} as a sequence of inner-loop iterations. Each iteration corresponds to one complete cycle of the system and contains the outputs of three roles: a \textsc{Reasoner} (proposal), a \textsc{Monitor} (audit), and a \textsc{Controller} (decision). The trace is designed to capture \emph{what information is produced and exchanged} during test-time evolution.

\paragraph{Overall structure.}
A trace is a chronological list of iteration records (\texttt{history}), ordered from the earliest to the latest:
\begin{quote}\small
\texttt{
history: [ \{ ... iteration record ... \}, \{ ... \}, \dots ]
}
\end{quote}

\paragraph{Iteration record.}
Each iteration record contains:
\begin{itemize}
  \item \textbf{Iteration index}: an integer indicating which inner-loop cycle this record corresponds to (1-indexed).
  \item \textbf{Reasoner output}: the candidate solution produced in this cycle.
  \item \textbf{Monitor output}: the diagnostic report produced by auditing the reasoner output.
  \item \textbf{Controller output}: the control decision that determines whether to accept, patch, or request a restart.
\end{itemize}

\paragraph{Reasoner output.}
The reasoner output contains:
\begin{itemize}
  \item \textbf{Final answer}: the extracted final answer for this iteration.
  \item \textbf{Full response text}: the complete natural-language solution text produced by the reasoner (including intermediate steps).
  \item \textbf{Updated role policy}: optional metadata describing a policy update applied to the reasoner for this iteration, including whether an update was used, whether it succeeded, which model performed the update, and the prompts before/after the update. (See Appendix~\ref{sec:appendix_prompt_update_impl} for how the updated prompts are composed.)
\end{itemize}

\paragraph{Monitor output.}
The monitor output contains:
\begin{itemize}
  \item \textbf{Error found flag}: a boolean indicating whether the monitor identified an error that could affect the final answer.
  \item \textbf{Error location}: an index or textual pointer indicating where the first major error was detected (or \texttt{NONE} if no error is found).
  \item \textbf{Error description}: a short actionable description of the detected issue.
  \item \textbf{Audit explanation}: a natural-language justification of the monitor's judgment.
  \item \textbf{Full report text}: the raw monitor report string, typically containing both the explanation and a structured summary.
  \item \textbf{Updated role policy}: optional metadata describing a policy update applied to the monitor for this iteration. (See Appendix~\ref{sec:appendix_prompt_update_impl}.)
\end{itemize}

\paragraph{Controller output.}
The controller output contains:
\begin{itemize}
  \item \textbf{Action}: a discrete decision code indicating how the system proceeds. We use three actions:
  \begin{itemize}
    \item \textbf{Accept}: accept the current iteration's answer and terminate the inner loop.
    \item \textbf{Patch}: provide a corrected reasoning (if available) and a corrected final answer, then terminate the inner loop.
    \item \textbf{Restart}: reject the current attempt and provide high-level revision suggestions for the next iteration.
  \end{itemize}
  For logging convenience, we encode actions as integers: \textsc{Accept}$=1$, \textsc{Patch}$=2$, \textsc{Restart}$=3$.
  \item \textbf{Justification}: a short explanation for why the controller selected the action.
  \item \textbf{Revision suggestions}: when the action is \emph{restart}, a brief action-oriented plan for the reasoner to attempt next.
  \item \textbf{Corrected reasoning}: when the action is \emph{patch}, the controller's corrected reasoning text.
  \item \textbf{Final answer}: when the action is \emph{accept} or \emph{patch}, the accepted/patched final answer; otherwise empty.
  \item \textbf{Full decision text}: the raw controller decision string, typically including both justification and a structured summary.
  \item \textbf{Updated role policy}: optional metadata describing a policy update applied to the controller for this iteration. (See Appendix~\ref{sec:appendix_prompt_update_impl}.)
\end{itemize}

\paragraph{Termination condition.}
An instance trace ends when the controller emits \emph{accept} or \emph{patch}, or when a maximum iteration budget is reached (in which case the last record is typically a \emph{restart}). The resulting \texttt{history} list thus provides a complete chronological record of proposals, audits, and control decisions for the instance.

\subsection{Retrieval Settings for In-batch Lesson Conditioning}
\label{sec:retrieval-settings}

For each sample $x_i$ in the next batch, we first retrieve up to $k$ most relevant lessons from the lesson buffer restricted to a sliding window of size $w$:
\begin{equation}
\label{eq:retrieve_impl}
\widetilde{\mathcal{L}}_{t}
=\mathrm{Retrieve}\!\left(x_i,\; \mathcal{L}_{b}^{(w)} \mid k\right),
\end{equation}
where $\mathcal{L}_{b}^{(w)}$ denotes the subset of buffered lessons that fall inside the most recent window of length $w$ (i.e., the last $w$ batches/steps, depending on the buffer implementation). If fewer than $k$ lessons are available in the window, we return all available lessons.

\paragraph{Similarity function.}
We perform retrieval by cosine similarity between the representation of the current problem and that of each candidate lesson.
We obtain embeddings using \textbf{text-embedding-qwen3-embedding-0.6b}.
Concretely, let $\mathbf{e}(x_i)$ be the embedding of the problem text $x_i$, and let $\mathbf{e}(\ell)$ be the embedding of a lesson $\ell\in\mathcal{L}_{b}^{(w)}$ (computed from the lesson text in its serialized form, i.e., concatenating its main fields such as \texttt{trigger} and \texttt{action}).
The cosine similarity is:
\begin{equation}
\label{eq:cosine}
s(x_i,\ell)=
\frac{\mathbf{e}(x_i)^\top \mathbf{e}(\ell)}
{\|\mathbf{e}(x_i)\|_2\,\|\mathbf{e}(\ell)\|_2}.
\end{equation}
The retrieval operator $\mathrm{Retrieve}(\cdot)$ ranks candidates in $\mathcal{L}_{b}^{(w)}$ by $s(x_i,\ell)$ and returns the top-$k$ lessons.

\paragraph{Hyperparameters.}
Unless otherwise stated, we use:
\begin{itemize}
  \item \textbf{Top-$k$ retrieval:} $k=3$.
  \item \textbf{Window size:} $w=3$.
  \item \textbf{Embedding model:} \textbf{text-embedding-qwen3-embedding-0.6b}.
  \item \textbf{Metric:} cosine similarity as in Eq.~\eqref{eq:cosine}.
\end{itemize}

\subsection{Key Hyperparameters and Runtime Budget}
\label{sec:hyperparams-budget}

\subsubsection{Batch Size}
\label{sec:batch-size}

We evaluate on four datasets: MATH500 (500 problems), GSM8K (1319 problems), TheoremQA (800 problems), and Game of 24 (100 problems).
For each dataset, the batch size is set to one-tenth of the dataset size, with a lower bound of 10 and an upper bound of 100.

Let $N$ denote the number of instances in a dataset. The batch size $B$ is computed as:
\begin{equation}
\label{eq:batch-size}
B \;=\; \min\!\left(100,\; \max\!\left(10,\; \left\lfloor \frac{N}{10} \right\rfloor \right)\right).
\end{equation}
The number of batches $M$ for a dataset is:
\begin{equation}
\label{eq:num-batches}
M \;=\; \left\lceil \frac{N}{B} \right\rceil.
\end{equation}
\begin{table}[t]
\centering
\caption{Batch size configuration and resulting number of batches for each dataset.}
\label{tab:batch-size}
\small
\setlength{\tabcolsep}{5pt}
\begin{tabular}{lrrr}
\toprule
Dataset & Dataset size $N$ & Batch size $B$ & \#Batches $M$ \\
\midrule
MATH500     & 500  & 50  & 10 \\
GSM8K       & 1319 & 100 & 14 \\
TheoremQA   & 800  & 80  & 10 \\
Game of 24  & 100  & 10  & 10 \\
\bottomrule
\end{tabular}
\end{table}

\subsubsection{Backbone Models}
\label{sec:backbone-models}

\paragraph{API-based models.}
\textbf{GPT-4o-mini} is decoded using the provider's default decoding parameters (i.e., we do not override sampling or penalty parameters).
\textbf{Gemini-2.0-flash} is also decoded with the provider's default generation configuration.

\paragraph{Open-source models.}
We evaluate two open-source backbones: \textbf{Qwen3-8B} and \textbf{Llama-3.1-8B-Instruct}.
For \textbf{Qwen3-8B}, the thinking mode is disabled; otherwise, decoding uses default settings (no additional manual tuning).
For \textbf{Llama-3.1-8B-Instruct}, all decoding parameters remain at default values.

\paragraph{No explicit output-token cap.}
Across \emph{all} models and methods, we do not impose an explicit upper bound on output tokens.
Generation terminates naturally (e.g., by end-of-sequence) or by provider-enforced limits.

\section{Iteration Trade-off Research}
\label{sec:appendix_iteration_tradeoff}

We study the iteration--cost trade-off by varying the maximum iteration budget in the inner loop. We run \texttt{gpt-4o-mini} on MATH500 using the same pipeline as our full method, with the \emph{only} difference being the setting of \texttt{maxiteration}. For each \texttt{maxiteration} value, we conduct one full evaluation and report accuracy (\%) and the average number of decoding tokens per instance.

\begin{table}[t]
\centering
\small
\caption{Effect of \texttt{maxiteration} on accuracy and average decoding tokens (MATH500, \texttt{gpt-4o-mini}).}
\label{tab:iter-tradeoff}
\begin{tabular}{ccc}
\toprule
\texttt{maxiteration} & Accuracy (\%) & \#Tokens \\
\midrule
2 & 84.6 & 1430.36 \\
3 & 85.8 & 1576.24 \\
4 & 83.2 & 1707.99 \\
5 & 82.4 & 1861.50 \\
6 & 84.4 & 1945.70 \\
\bottomrule
\end{tabular}
\end{table}

\begin{figure}[t]
  \centering
  \includegraphics[width=0.88\linewidth]{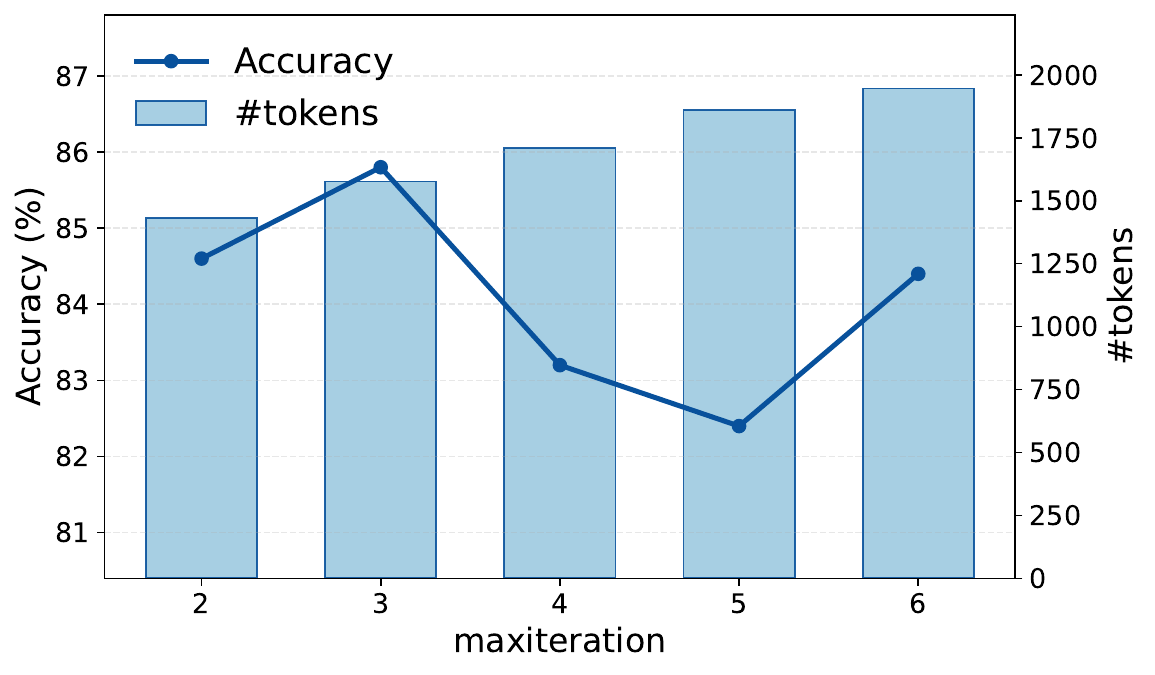}
  \caption{Iteration budget trade-off on MATH500 with \texttt{gpt-4o-mini}: accuracy (line) versus average decoding tokens (bars) under different \texttt{maxiteration}.}
  \label{fig:iter-tradeoff}
\end{figure}

\paragraph{Discussion.}
A smaller \texttt{maxiteration} can be \emph{too short} for hard instances. With limited inner-loop cycles, the system may terminate before the reasoner has enough opportunities to incorporate monitor feedback and controller guidance. As a result, correctable mistakes may remain unresolved (or the controller may be forced into premature restart/accept decisions), typically reducing accuracy even though fewer iterations save tokens.

Conversely, a larger \texttt{maxiteration} can be \emph{too long}. Increasing the iteration budget monotonically increases generation cost because each additional cycle incurs extra proposals, audits, and decisions. In addition, allowing many iterations can introduce over-refinement effects: later cycles may over-correct a nearly-correct solution, drift away from a good partial derivation, or accumulate inconsistencies across successive revisions. This can produce diminishing (or even negative) returns in accuracy while token usage continues to rise.

Given these trade-offs, using a small number of iterations is desirable for efficiency, but it must remain sufficient for feedback-driven correction. In our setting, \texttt{maxiteration}=3 offers a reasonable balance: it enables an extra round of revision beyond a short budget, while avoiding the heavier cost and potential over-refinement associated with longer iteration limits.

\section[Does the MC-squared role-strategy update inject task-relevant Reasoner guidance?]{Does \textsc{MC$^2$}'s role-strategy update inject task-relevant Reasoner guidance?}
\label{app:guidance-relevance}

This experiment evaluates whether the \emph{role-strategy update} in \textsc{MC$^2$} (i.e., updating the Reasoner policy prompt via retrieved micro-lessons and consolidated meta-knowledge; cf. the \emph{update MRO role strategy} step) injects \emph{task-relevant} guidance for the current instance. Importantly, we aim to measure relevance beyond trivial prompt overlap with the problem statement.

\paragraph{Setup.}
We evaluate on \textbf{MATH500} using the \textbf{GPT-4o-mini} backbone. Following Section~\ref{sec:batch-size}, the batch size is $B{=}50$ (thus $M{=}10$ batches). For each instance $i$ in batch $b$, we take the Reasoner's \emph{compiled} prompt after the role-strategy update (the rewritten/augmented prompt used to call the Reasoner) and explicitly remove the full problem text from it, yielding a \emph{guidance-only} snippet:
\[
g_i \;=\; \texttt{FinalPrompt}_i \;-\; \texttt{Question}_i.
\]
We then compute the cosine similarity between the instance question and this guidance-only snippet using the same embedding model as in our retrieval component:
\[
s_i \;=\; \cos\!\big(\text{Emb}(\texttt{Question}_i),\;\text{Emb}(g_i)\big).
\]
As a within-batch control baseline, we randomly sample a guidance-only snippet $g_{j}$ from \emph{the same batch} ($j \neq i$) and compute:
\[
s_i^{\text{rand}} \;=\; \cos\!\big(\text{Emb}(\texttt{Question}_i),\;\text{Emb}(g_{j})\big).
\]
For multi-turn instances, we compute similarities per turn and report their mean. Finally, for each batch $b$, we report the mean similarity over instances in that batch. Because batch~1 has no prior accumulated experience to retrieve/compile into the role strategy, we report batches~2--10.

\paragraph{What this shows.}
Because the question text is explicitly removed from the updated prompt, high similarity cannot be explained by copying or restating the problem. A consistently higher \texttt{reasonerguide} similarity than the within-batch \texttt{random guide} baseline indicates that the \emph{role-strategy update} injects \emph{instance-aligned} guidance (i.e., cues that are specific to the current problem) rather than generic advice. The stable gap across batches further supports that the \textsc{MC$^2$} update mechanism reliably compiles accumulated experience into targeted Reasoner guidance as meta-knowledge grows.

\paragraph{LLM-as-judge corroboration.}
As a complementary check, we use \textbf{GPT-5} as a binary judge to evaluate whether the updated role policies contains \emph{task-specific key operations/checkpoints} that would materially help solve the given question. The judge is prompted with the instance question and the agent's updated role policies, and outputs a strict binary decision per turn; for multi-turn instances we average across turns and then report the per-agent mean.

\begin{figure}[t]
\centering

\begin{minipage}[t]{0.98\columnwidth}
\centering
\small
\setlength{\tabcolsep}{8pt}
\begin{tabular}[t]{lc}
\toprule
\textbf{Agent} & \textbf{$p_{\text{keyops}}$} \\
\midrule
Reasoner   & 0.9146 \\
Monitor    & 0.9618 \\
Controller & 0.9848 \\
\bottomrule
\end{tabular}
\captionsetup{hypcap=false}
\captionof{table}{GPT-5 binary-judge evaluation of whether the agent's updated role policies contains task-specific key operations/checkpoints that would materially help solve the given question. Results are averaged per instance (and across turns for multi-turn instances), then averaged over instances.}
\label{tab:keyops-rate}
\end{minipage}

\vspace{0.8em}

\begin{minipage}[t]{0.98\columnwidth}
\centering
\begin{tikzpicture}
\begin{axis}[
    width=0.9\linewidth,
    height=0.42\linewidth,
    xlabel={Batch index $b$},
    ylabel={Cosine similarity},
    xmin=2, xmax=10,
    ymin=0.15, ymax=0.80,
    grid=both,
    legend style={at={(0.02,0.98)},anchor=north west,draw=none,fill=none},
    tick label style={font=\small},
    label style={font=\small},
]
\addplot+[mark=*] coordinates {
    (2,0.724116701)
    (3,0.723292754)
    (4,0.736890826)
    (5,0.730193113)
    (6,0.732046205)
    (7,0.72785623)
    (8,0.721366348)
    (9,0.734951086)
    (10,0.719628982)
};
\addlegendentry{\texttt{reasonerguide}}

\addplot+[mark=square*] coordinates {
    (2,0.251700578)
    (3,0.248236282)
    (4,0.254713941)
    (5,0.247755928)
    (6,0.222861681)
    (7,0.252350184)
    (8,0.228775744)
    (9,0.262090672)
    (10,0.26546875)
};
\addlegendentry{\texttt{random guide}}
\end{axis}
\end{tikzpicture}
\caption{Embedding cosine similarity between each question and the \emph{guidance-only} snippet produced by the \textsc{MC$^2$} role-strategy update for the Reasoner (question text removed), compared with a within-batch random baseline. Results are averaged per batch ($B{=}50$) on MATH500 using GPT-4o-mini.}
\label{fig:guidance-relevance}
\end{minipage}

\vspace{-0.6em}
\end{figure}

\section{Does data order affect the consolidated global meta-knowledge?}
\label{sec:appendix_order_sensitivity}

Our global meta-knowledge is updated via a bounded sliding-window consolidation mechanism (Eq.~\eqref{eq:meta}), which raises a natural question: \emph{does the presentation order of input instances affect the resulting meta-knowledge and final performance?}
To test order sensitivity, we run MC$^2$ with two API backbones, \texttt{gpt-4o-mini} and \texttt{gemini-2.0-flash}, on two datasets, \textsc{Game of 24} and \textsc{MATH500}.
We compare the dataset's released instance order (i.e., the original file order used in our data loader) with a randomly permuted order (shuffled instances) under the same hyperparameters and evaluation protocol.
For the shuffled setting, each run uses a different random seed for permutation.
Overall, we observe that shuffling the dataset order has negligible impact on final performance, suggesting that the sliding-window consolidation is reasonably robust to instance ordering.

\begin{table}[t]
\centering
\small
\caption{Order sensitivity under sliding-window meta-knowledge consolidation. $\Delta$ is computed as \texttt{Shuffled} $-$ \texttt{Default} and shown inline in the shuffled column.}
\label{tab:order_sensitivity}
\setlength{\tabcolsep}{12pt}
\renewcommand{\arraystretch}{1.15}
\begin{tabular}{lcc}
\toprule
 & \textbf{Default order} & \textbf{Shuffled order}\\
\midrule
 & \multicolumn{2}{c}{\textbf{gpt-4o-mini}} \\
math500    & 85.13 & 85.06 \,(\,$-0.07$\,) \\
game-of-24 & 88.67 & 87.33 \,(\,$-1.34$\,) \\
\midrule
 & \multicolumn{2}{c}{\textbf{gemini-2.0-flash}} \\
math500    & 96.73 & 96.93 \,(+\,$0.20$\,) \\
game-of-24 & 94.33 & 93.67 \,(\,$-0.66$\,) \\
\bottomrule
\end{tabular}
\end{table}

\section{How transferable is consolidated meta-knowledge across tasks?}
\label{sec:appendix_cross_task_transfer}

MC$^2$ is currently designed primarily for within-task learning. We start from single-task settings to validate the core premise of the framework, namely that the system can achieve self-improvement by consolidating lessons from its own experience. Cross-task generalization is therefore not a main claim of the current paper, but it is still useful to ask whether the consolidated meta-knowledge can provide a helpful initialization when transferred across related benchmarks.

To study this question, we conduct a preliminary cross-task transferability analysis. In \textbf{Transfer Warm-Start}, the consolidated meta-knowledge learned on a source task is used to initialize the target task, after which MC$^2$ continues running on the target task. We compare this setting against \textbf{Full MC$^2$ (Within-Task)}, where the full MC$^2$ pipeline is run directly on the target task using only target-task experience.

\begin{table}[H]
\centering
\small
\caption{Preliminary cross-task transferability of consolidated meta-knowledge.}
\label{tab:cross_task_transfer}
\setlength{\tabcolsep}{5pt}
\renewcommand{\arraystretch}{1.12}
\resizebox{\columnwidth}{!}{%
\begin{tabular}{@{}lcc@{}}
\toprule
\textbf{Source $\rightarrow$ Target} & \textbf{Transfer Warm-Start} & \textbf{Full MC$^2$ (Within-Task)} \\
\midrule
GSM8K $\rightarrow$ MATH-500     & 83.60 & 85.13 \\
GSM8K $\rightarrow$ TheoremQA    & 57.00 & 55.96 \\
MATH-500 $\rightarrow$ GSM8K     & 96.36 & 96.92 \\
MATH-500 $\rightarrow$ TheoremQA & 54.63 & 55.96 \\
\bottomrule
\end{tabular}
}
\end{table}

The results suggest that cross-task transfer is feasible, but clearly direction-dependent. In particular, transferring from MATH-500 to GSM8K preserves almost all of the within-task performance (96.36\% vs.\ 96.92\%), while GSM8K $\rightarrow$ TheoremQA remains close to the within-task result (57.00\% vs.\ 55.96\%), which may indicate that some arithmetic checking and verification habits remain useful across tasks. Other transfer directions show small drops relative to within-task consolidation. This likely reflects a combination of two factors: part of the consolidated meta-knowledge is task-specific, and part of the observed gap may come from dataset-level noise. Overall, these results provide preliminary evidence of cross-task transferability, while also confirming that the strongest and most reliable gains currently come from within-task consolidation.

\section{Example}
\vspace{-0.2em}

We first visualize one concrete \textsc{MC$^2$} episode as four single-column cards.

\subsection{Example Cards}

\definecolor{CardBlue}{RGB}{232,243,255}
\definecolor{CardGreen}{RGB}{235,250,242}
\definecolor{CardYellow}{RGB}{255,249,230}
\definecolor{CardPink}{RGB}{255,238,244}

\tcbset{
  mycard/.style={
    enhanced,
    unbreakable,
    sharp corners,
    boxrule=0.6pt,
    colframe=black!35,
    fontupper=\small,
    before upper=\raggedright,
    left=6pt,right=6pt,top=5pt,bottom=5pt,
    before skip=4pt, after skip=6pt,
    width=\linewidth
  }
}

\newcommand{\CardField}[2]{\textbf{#1}: #2\par}
\newcommand{\CardMono}[1]{\texttt{#1}}
\newcommand{\CardHeader}[1]{%
  {\bfseries\small #1\par}%
  \vspace{2pt}%
  \hrule height 0.4pt\relax
  \vspace{5pt}%
}

\newcommand{\CardFigure}[5]{%
  \par\noindent
  \begin{minipage}{\linewidth}
    \centering
    \begin{tcolorbox}[mycard={#1}, colback=#2]
      \CardHeader{#1}
      #5
    \end{tcolorbox}
    \captionsetup{hypcap=false}
    \captionof{figure}{#3}\label{#4}
  \end{minipage}
  \par
}

\CardFigure
  {Card 1: Instance Overview}
  {CardBlue}
  {Card 1: Instance overview (Index 400).}
  {fig:mrmc-card1}
  {%
    \CardField{Index}{400}
    \CardField{Subject}{Geometry}
    \CardField{Question}{In the circle with center $Q$, radii $AQ$ and $BQ$ form a right angle. The two smaller regions are tangent semicircles. The radius of the circle with center $Q$ is 14 inches. Find the radius of the smaller semicircle.}
    \CardField{Gold answer}{$\dfrac{14}{3}$}
    \CardField{Model final}{$\tfrac{14}{3}$}
    \CardField{Status}{accepted}
    \CardField{Iterations}{2}
    \CardField{Rank}{\CardMono{task\_quality=B; task\_outcome=success; reasoner=R\_good; monitor=M\_ok; controller=C\_good}}
  }

\CardFigure
  {Card 2: Trajectory at a Glance}
  {CardGreen}
  {Card 2: MRO trajectory at a glance.}
  {fig:mrmc-card2}
  {%
    \textbf{Iter 1}\par
    \begin{itemize}[leftmargin=*, nosep]
      \item \textbf{Reasoner answer}: $7$
      \item \textbf{Monitor}: \CardMono{error\_found=YES}, \CardMono{error\_step=4}
      \item \textbf{Monitor note}: The equation $2r=28$ is incorrect (segment relationship is mis-modeled).
      \item \textbf{Controller}: \CardMono{action=RESTART (Action 3)}
      \item \textbf{Justification}: Major geometric relationship issues; restart with correct configuration.
      \item \textbf{Outcome}: continue
    \end{itemize}

    \tcbline

    \textbf{Iter 2}\par
    \begin{itemize}[leftmargin=*, nosep]
      \item \textbf{Reasoner answer}: $\tfrac{14}{3}$
      \item \textbf{Monitor}: \CardMono{error\_found=NO}
      \item \textbf{Monitor note}: All steps are consistent and correct.
      \item \textbf{Controller}: \CardMono{action=ACCEPT (Action 1)}
      \item \textbf{Outcome}: terminate
    \end{itemize}
  }

\CardFigure
  {Card 3: Iteration 1 (Failure)}
  {CardYellow}
  {Card 3: Failure iteration and diagnosis.}
  {fig:mrmc-card3}
  {%
    \CardField{Reasoner final\_answer}{$7$}
    \CardField{Core issue (summary)}{Unjustified segment equality assumptions lead to an invalid equation for $r$.}
    \CardField{Monitor}{\CardMono{error\_found=YES}, \CardMono{error\_step=4}}
    \CardField{Monitor description}{The equation $2r=28$ neglects the actual placement/tangency constraints of the semicircles.}
    \CardField{Controller decision}{RESTART}
    \CardField{Controller suggestion}{Redefine centers and use tangency as a distance constraint (center distance = sum of radii).}
  }

\CardFigure
  {Card 4: Iteration 2 (Success)}
  {CardPink}
  {Card 4: Successful iteration and solution.}
  {fig:mrmc-card4}
  {%
    \textbf{Setup}\par
    $Q=(0,0)$, $A=(14,0)$, $B=(0,14)$. \par
    Semicircle on diameter $AQ$: radius $7$, center $C=(7,0)$. \par
    Smaller semicircle radius $r$, internally tangent to the big circle: center $D=(0,14-r)$. \par
    \vspace{2pt}
    \textbf{Tangency constraint (two semicircles tangent)}\par
    \[
      \sqrt{(7-0)^2 + (0-(14-r))^2} = 7+r.
    \]
    \textbf{Solve}\par
    \begin{align*}
      49 + (14-r)^2 &= (7+r)^2 \\
      196 &= 42r \\
      r &= \dfrac{14}{3}.
    \end{align*}

    \CardField{Final answer}{$\boxed{\tfrac{14}{3}}$}
  }

The Geometry episode (Index 400) completes in two MRO inner iterations: the first attempt outputs an incorrect
answer ($7$) due to a mis-modeled segment relationship; the monitor flags the error (error\_step=4) and the
controller triggers a restart; the second attempt uses coordinate placement and a center-distance tangency
equation to obtain the correct radius $\tfrac{14}{3}$.

\subsection{Representative Meta-Knowledge Across Benchmarks}
\label{sec:appendix_meta_knowledge_content}

Appendix~\ref{app:guidance-relevance} already shows that the updated role policies in \textsc{MC$^2$} contain task-relevant operations rather than generic advice. We now inspect the content of representative distilled micro-lessons directly. The examples below illustrate that the learned knowledge is \emph{reusable prompt-level meta-knowledge}---namely reasoning heuristics, verification routines, and control policies---rather than task answers or memorized solutions.

\begin{center}
\scriptsize
\captionsetup{hypcap=false}
\captionof{table}{Representative distilled meta-knowledge across benchmarks.}
\label{tab:appendix_meta_knowledge_content}
\setlength{\tabcolsep}{3pt}
\renewcommand{\arraystretch}{1.12}
\begin{tabularx}{\columnwidth}{>{\raggedright\arraybackslash}p{0.25\columnwidth}>{\raggedright\arraybackslash}X>{\raggedright\arraybackslash}p{0.19\columnwidth}}
\toprule
\textbf{Source Problem (Domain)} & \textbf{Distilled Micro-Lesson Content} & \textbf{Knowledge Type} \\
\midrule
\textbf{Algebraic sequence}:
$a_{i+1}=\frac{1}{1-a_i}$ with $a_3=a_1$ (MATH-500)
&
\textbf{Success}; tags: process, checking. \textbf{DO}: validate each algebraic transformation before committing to the final answer. \textbf{AVOID}: relying on shortcuts that skip a concrete verification step. Reflection anchor: insert a final derivation check before submission.
&
Iterative Verification\newline Heuristic
\\
\midrule
\textbf{Complex arithmetic}:
$w=\frac{3z+1}{5z+7}$ with $z=1+i$ (MATH-500)
&
\textbf{Success}; tags: arithmetic, accuracy. \textbf{DO}: re-check complex simplification and modulus calculations before finalizing. \textbf{AVOID}: ignoring monitor feedback on intermediate arithmetic. Reflection anchor: add a dedicated numeric check before the final answer.
&
Arithmetic Precision\newline Control
\\
\midrule
\textbf{Combinatorics}:
ordered partitions of 8 elements into 5 non-empty subsets (TheoremQA)
&
\textbf{Failure}; tags: logic, validation. \textbf{DO}: preserve an explicit counting structure and justify each combinatorial choice sequentially. \textbf{AVOID}: relying on unverified assumptions. Reflection anchor: name the first concrete counting mistake and why it matters.
&
Logical Structure\newline Validation
\\
\midrule
\textbf{Complex analysis}:
contour integral with $\sin(1/z)$ on $|z|=1$ (TheoremQA)
&
\textbf{Success}; tags: complex-analysis, integration. \textbf{DO}: use local series expansions to isolate the residue-bearing term. \textbf{AVOID}: discarding higher-order terms too early. Reflection anchor: the monitor confirms the Taylor-series route before final extraction.
&
Domain-Specific Tactic\newline (Residue Theorem Application)
\\
\midrule
\textbf{Calculus}:
$g=f^{-1},\ f(x)=x+\cos(x),\ g'(1)$ (TheoremQA)
&
\textbf{Success}; tags: clarity, logic. \textbf{DO}: keep each symbolic dependency explicit when differentiating inverse relations. \textbf{AVOID}: skipping intermediate transformations that obscure why the derivative expression is valid. Reflection anchor: preserve a stepwise trace that the monitor can audit.
&
Step-wise Transparency\newline Heuristic
\\
\bottomrule
\end{tabularx}
\end{center}

These examples reveal that the consolidated meta-knowledge spans three functional families. First, \textsc{MATH-500} tends to yield process-level verification and arithmetic-control rules, which emphasize explicit checking before answer commitment. Second, \textsc{TheoremQA} contributes more logical-structural and domain-specific tactical rules, such as preserving combinatorial structure or using residue-oriented series expansions. Third, the learned content is not a cache of task answers: successes contribute reusable \textbf{DO} rules, while failures contribute reusable \textbf{AVOID} constraints. This is exactly the form of prompt-level meta-knowledge produced by the distillation operator in Eq.~\eqref{eq:micro} and then stabilized by temporal consolidation in Eq.~\eqref{eq:meta}.

\section{Prompt}
\label{sec:appendix_prompt}

We provide the full prompt templates used by the Reasoner, Monitor, and Controller in MC$^2$, covering cold-start inference, inference with retrieved micro-lessons/meta-knowledge, role-policy updates, reflection-to-micro-lesson distillation, and windowed meta-knowledge consolidation.

\begingroup
\setlength{\textfloatsep}{6pt plus 2pt minus 2pt}
\setlength{\floatsep}{6pt plus 2pt minus 2pt}
\setlength{\intextsep}{6pt plus 2pt minus 2pt}
\setlength{\dbltextfloatsep}{6pt plus 2pt minus 2pt}
\setlength{\dblfloatsep}{6pt plus 2pt minus 2pt}
\setlength{\abovecaptionskip}{2pt}
\setlength{\belowcaptionskip}{0pt}

\makeatletter
\setlength{\@dblfptop}{0pt}
\setlength{\@dblfpsep}{6pt}
\setlength{\@dblfpbot}{0pt}
\setlength{\@fptop}{0pt}
\setlength{\@fpsep}{6pt}
\setlength{\@fpbot}{0pt}
\makeatother

\newcommand{\promptfig}[3]{%
\begin{figure*}[t]
    \centering
    \vspace{-2pt}
    \includegraphics[width=\textwidth,height=0.40\textheight,keepaspectratio]{#1}
    \vspace{-4pt}
    \caption{#2}
    \vspace{-6pt}
    \label{#3}
\end{figure*}
}

\promptfig{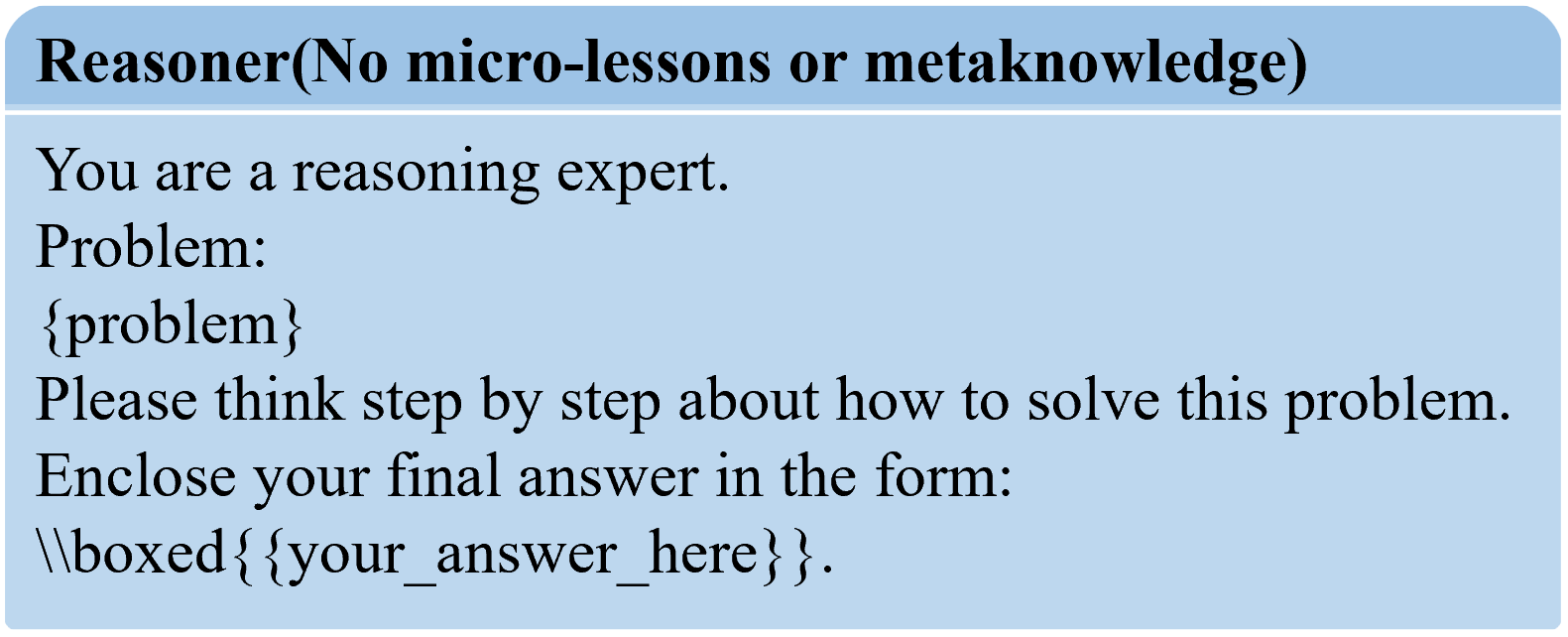}
{Reasoner prompt (cold start; no micro-lessons or meta-knowledge).}
{fig:prompt_reasoner_coldstart}

\promptfig{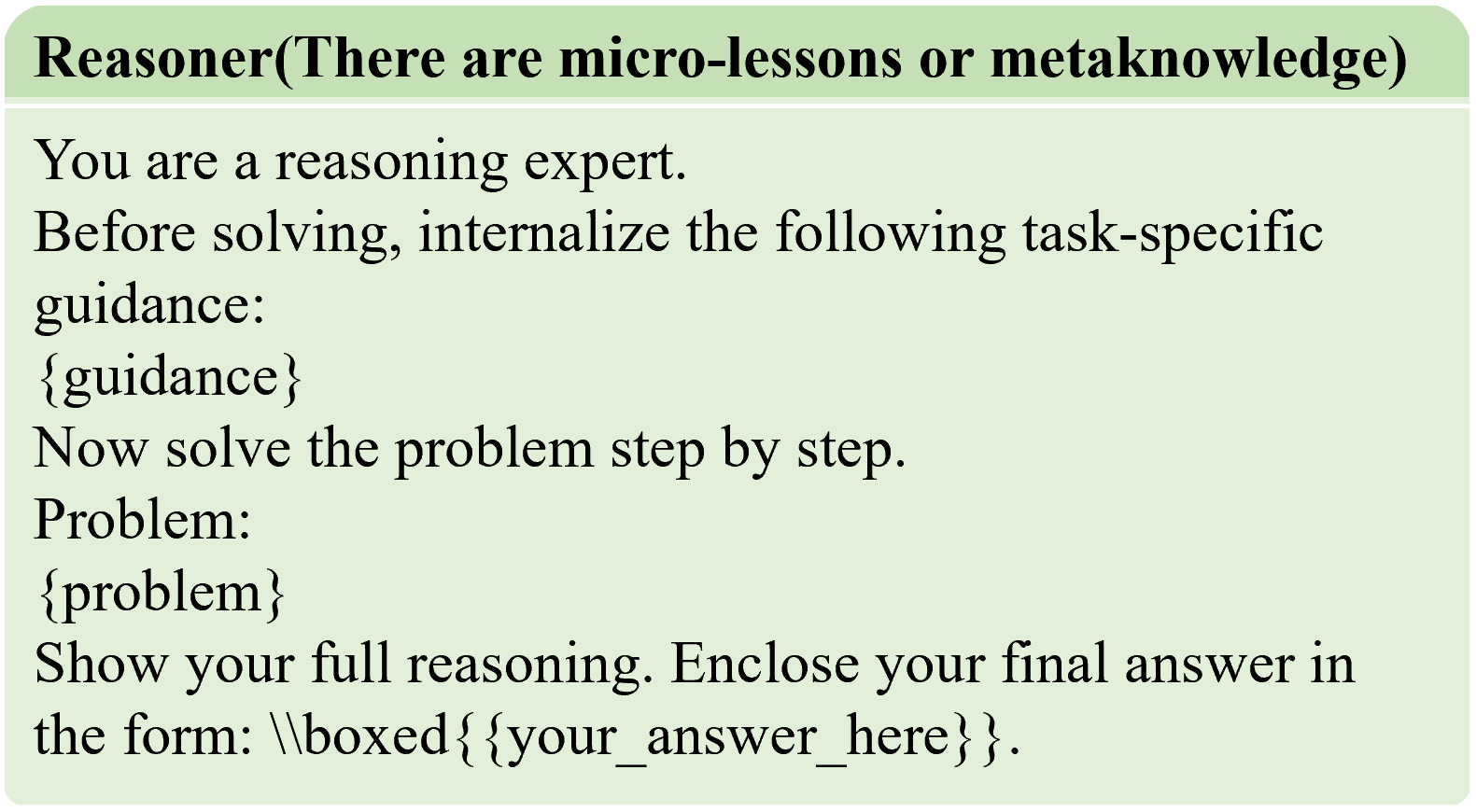}
{Reasoner prompt (with retrieved micro-lessons / meta-knowledge guidance).}
{fig:prompt_reasoner_guided}

\promptfig{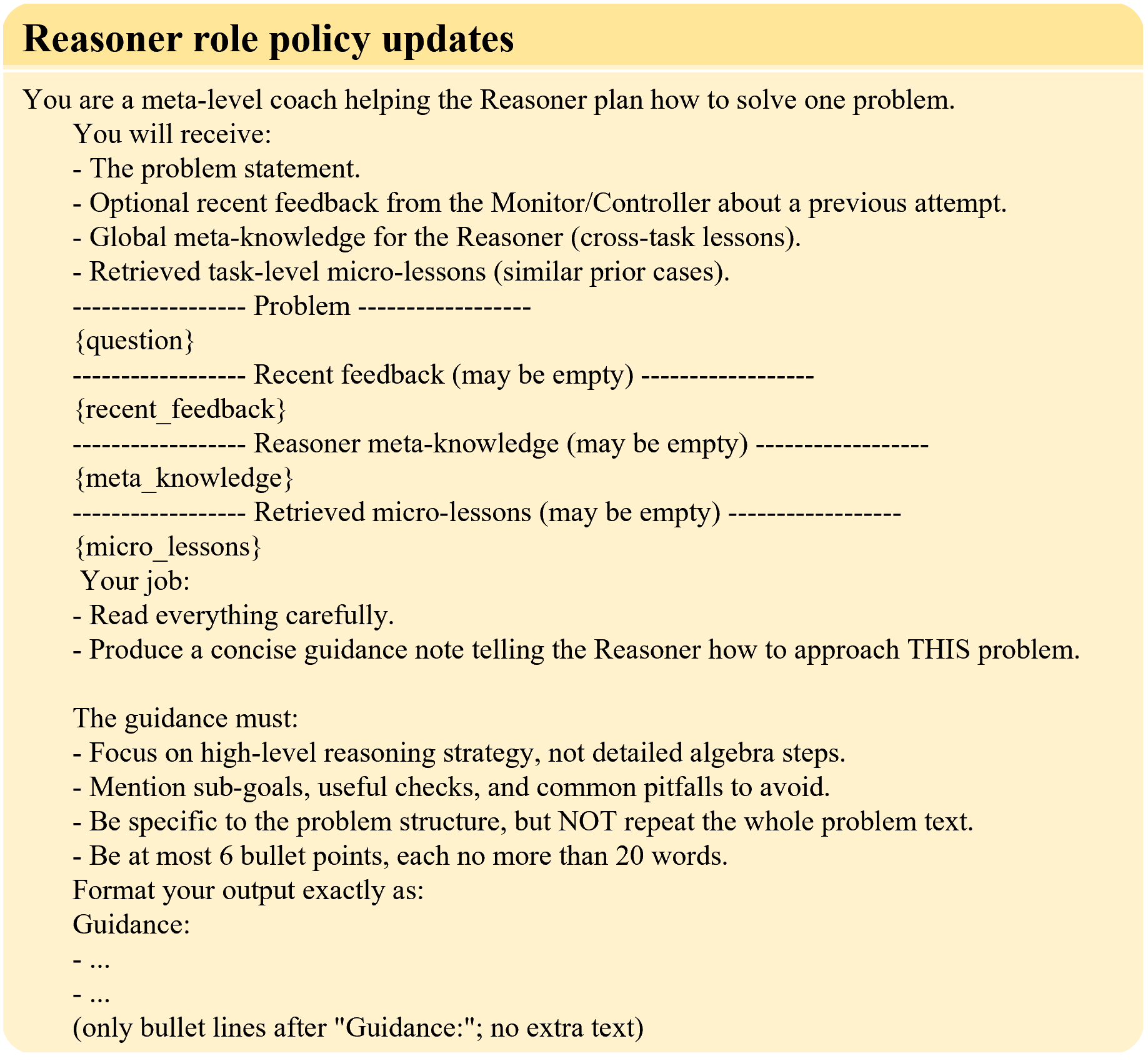}
{Reasoner role policy update prompt (meta-level coaching to generate task-specific guidance).}
{fig:prompt_reasoner_policy_update}

\promptfig{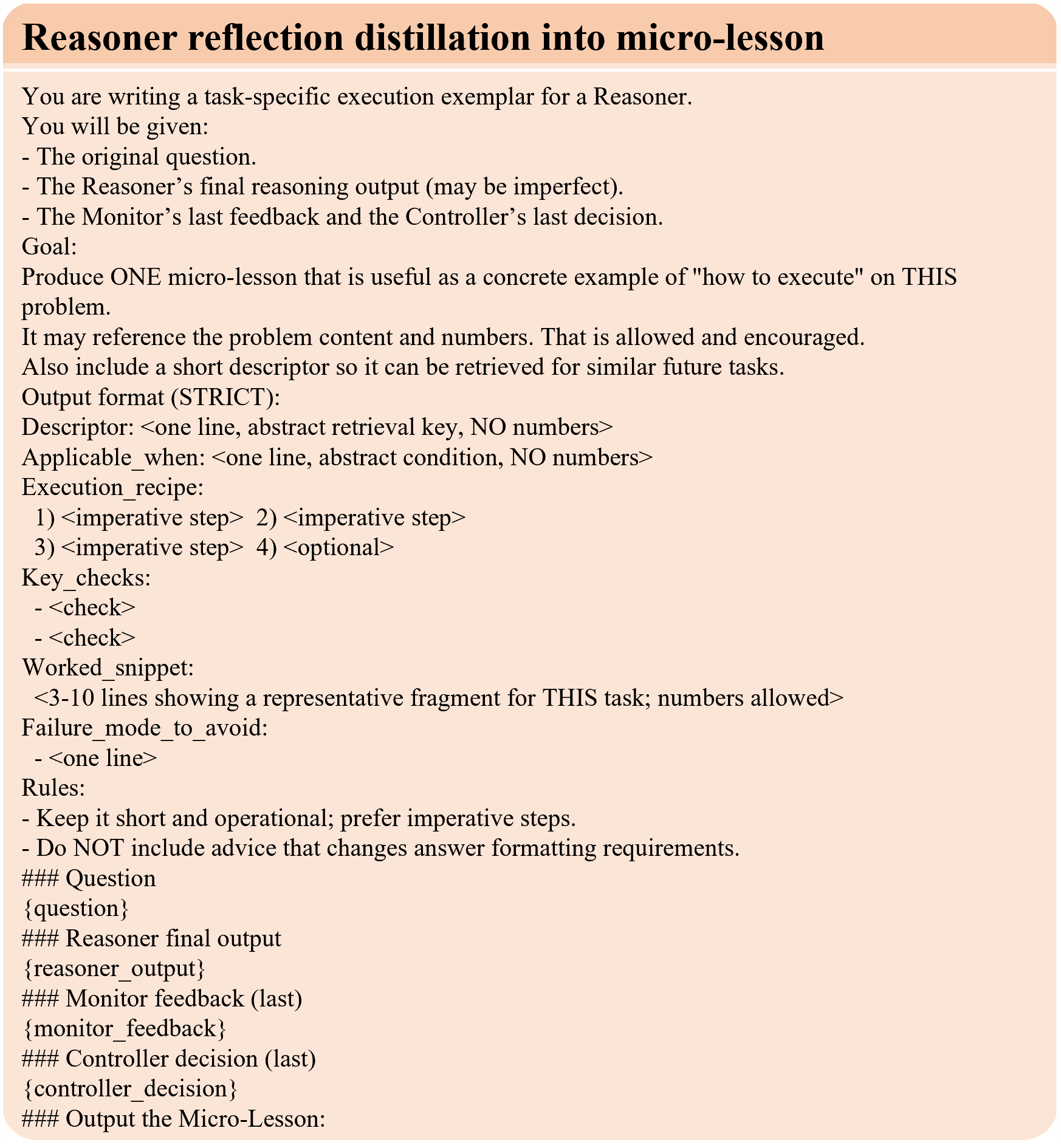}
{Reasoner reflection distillation prompt for micro-lesson (execution exemplar with checks and failure modes).}
{fig:prompt_reasoner_distill}

\promptfig{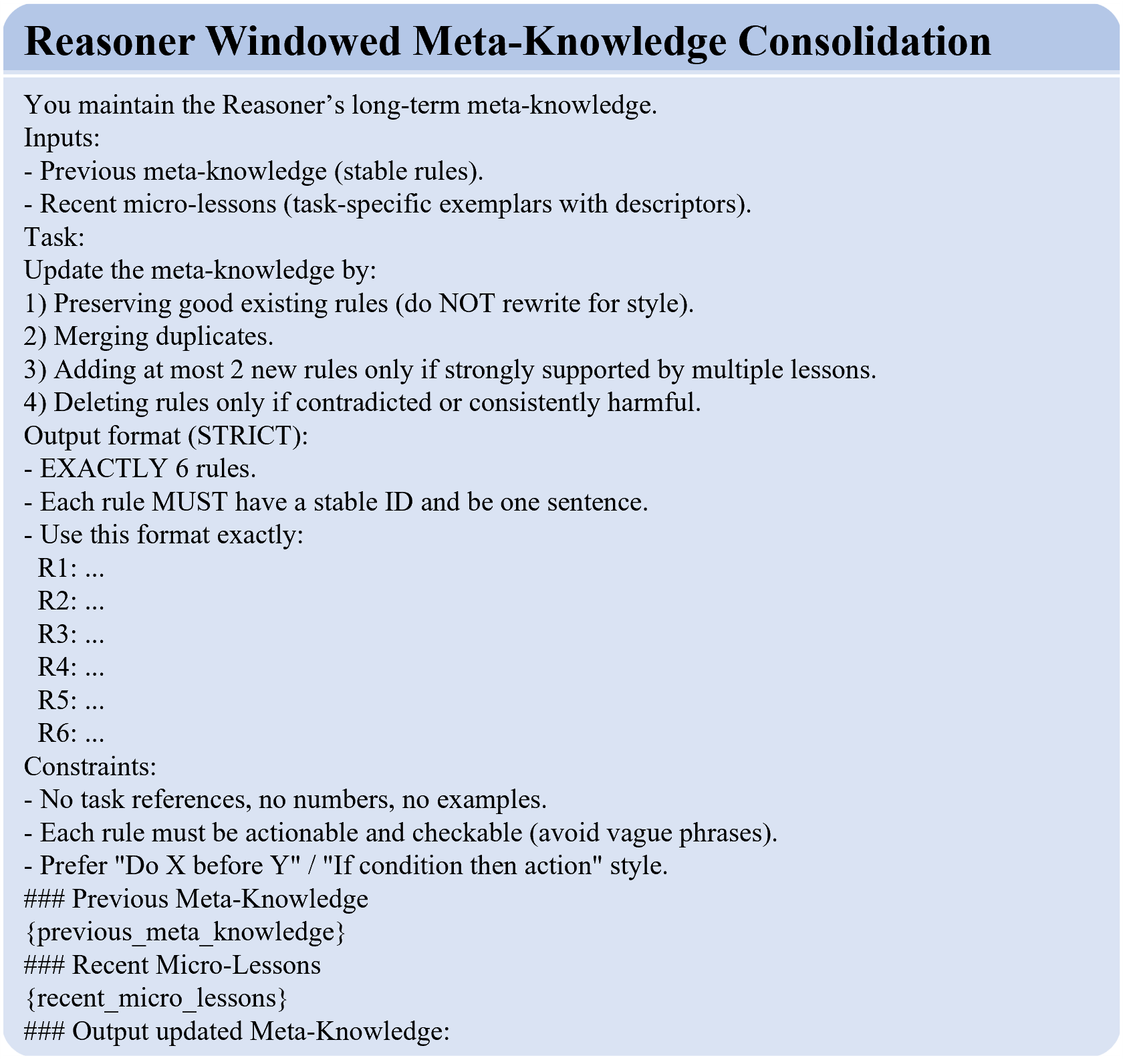}
{Reasoner prompt for windowed meta-knowledge consolidation (merge and update stable rules from recent micro-lessons).}
{fig:prompt_reasoner_consolidate}

\promptfig{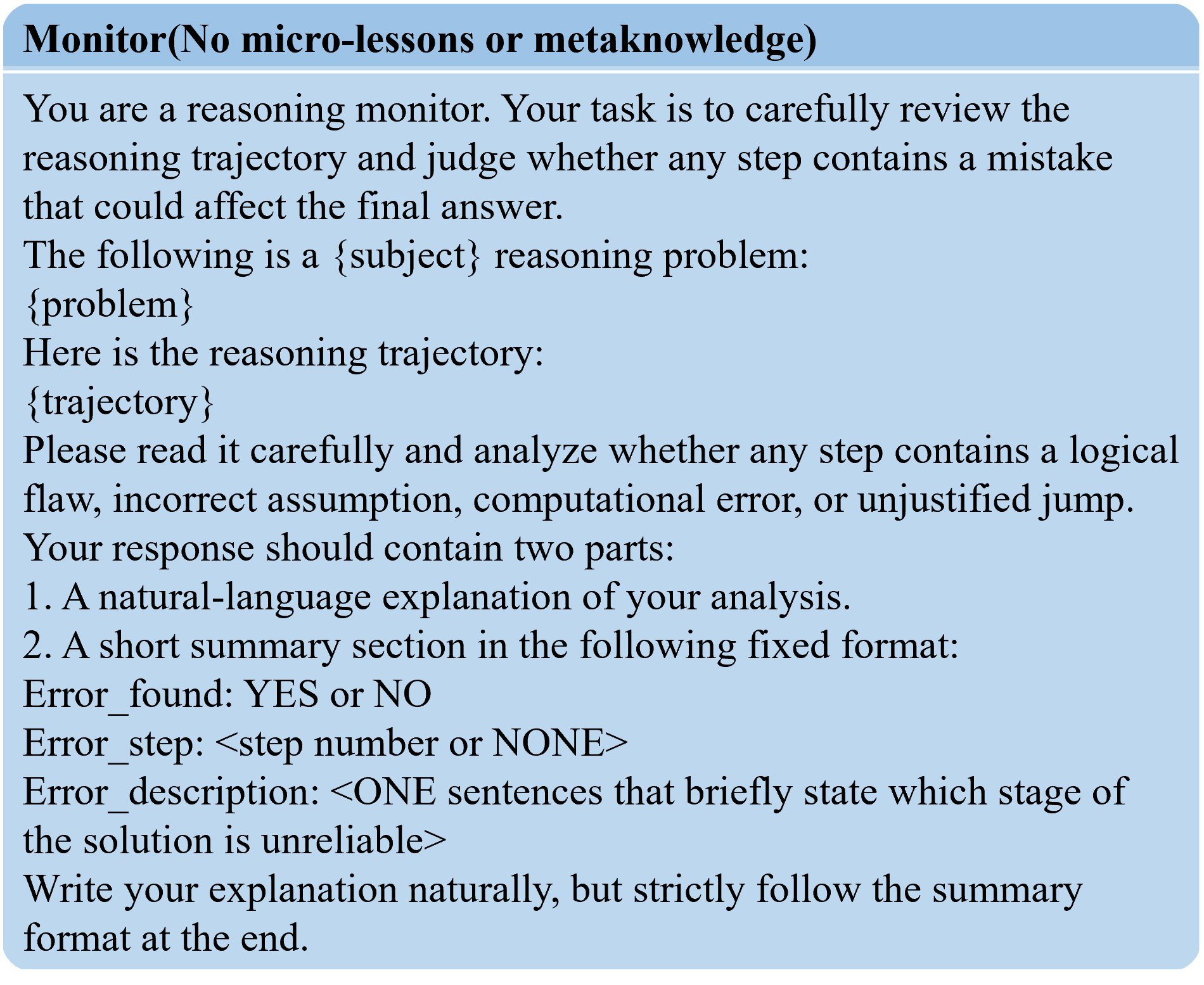}
{Monitor prompt (cold start; no micro-lessons or meta-knowledge).}
{fig:prompt_monitor_coldstart}

\promptfig{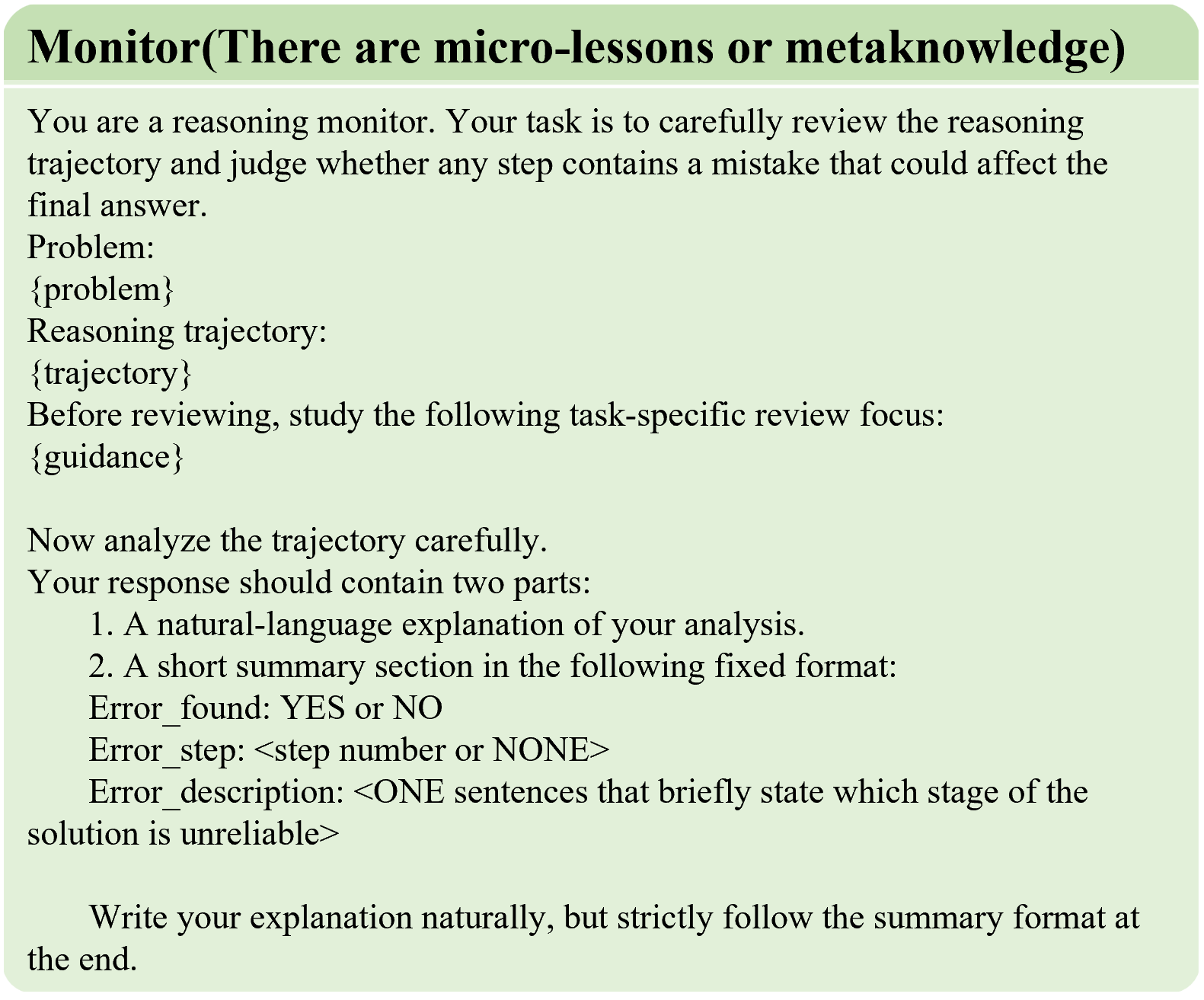}
{Monitor prompt (with retrieved micro-lessons / meta-knowledge review focus).}
{fig:prompt_monitor_guided}

\promptfig{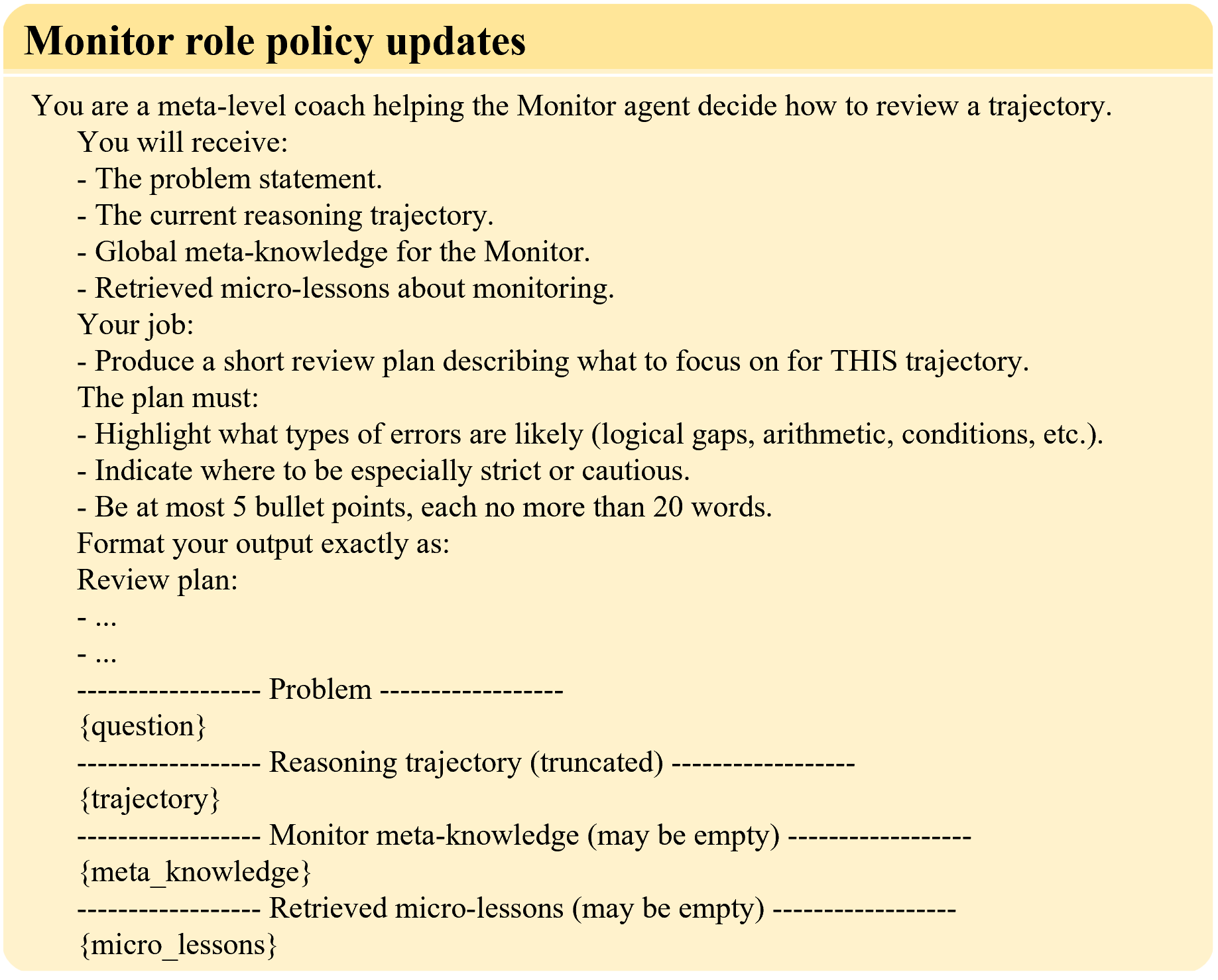}
{Monitor role policy update prompt (meta-level coaching to produce a concise review plan).}
{fig:prompt_monitor_policy_update}

\promptfig{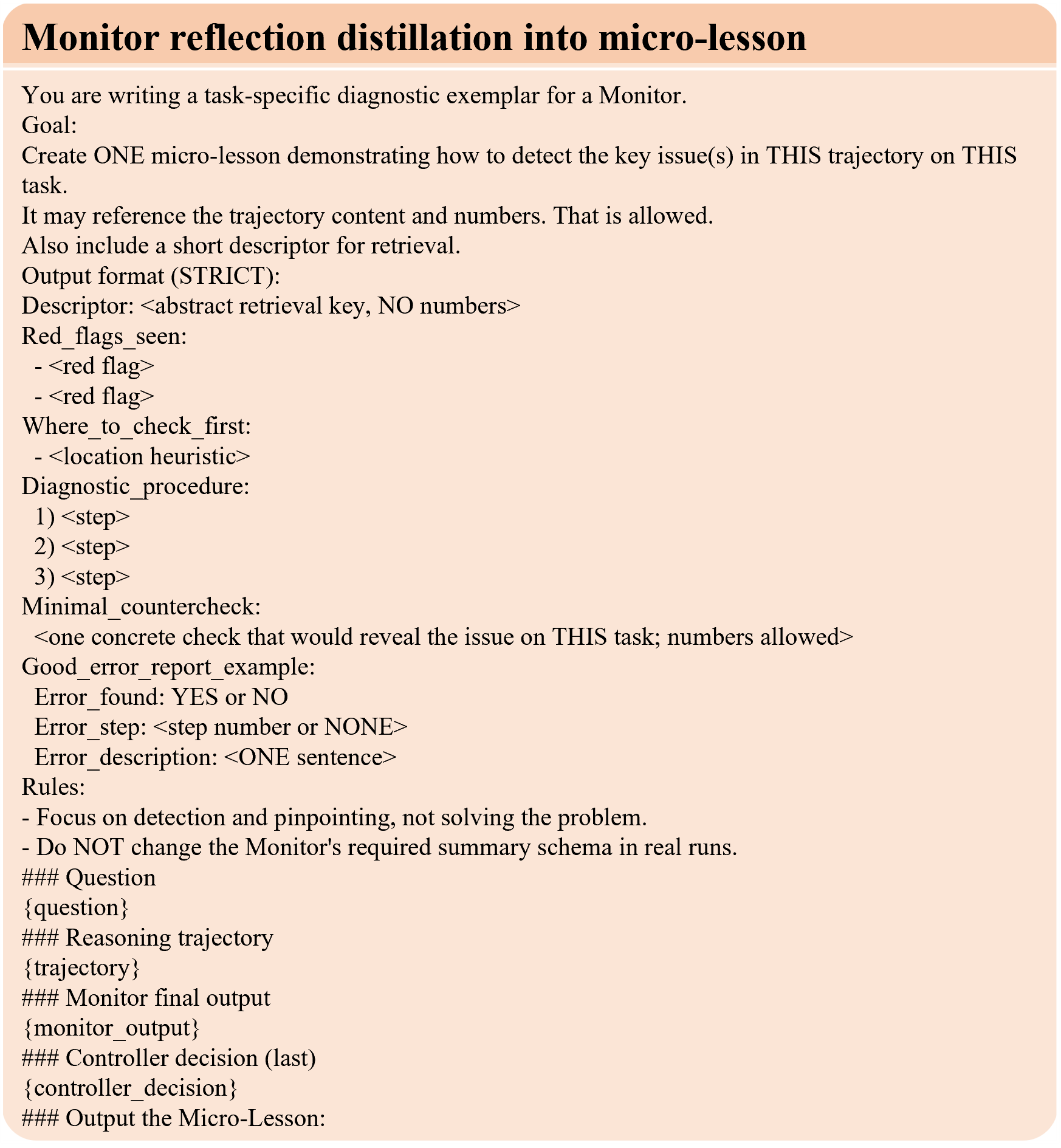}
{Monitor reflection distillation prompt for micro-lesson (diagnostic exemplar with red flags and minimal countercheck).}
{fig:prompt_monitor_distill}

\promptfig{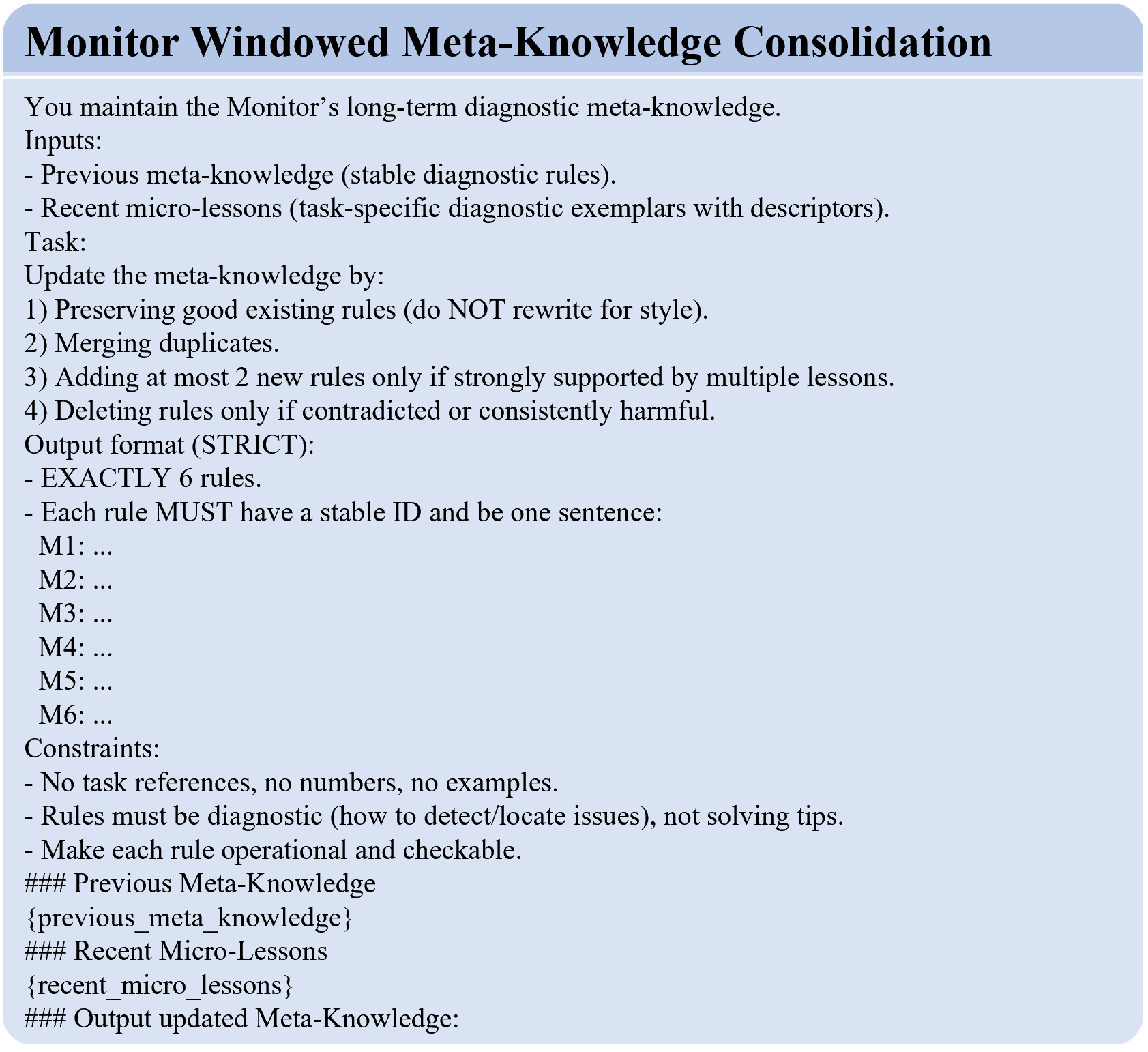}
{Monitor prompt for windowed meta-knowledge consolidation (update long-term diagnostic rules from recent micro-lessons).}
{fig:prompt_monitor_consolidate}

\promptfig{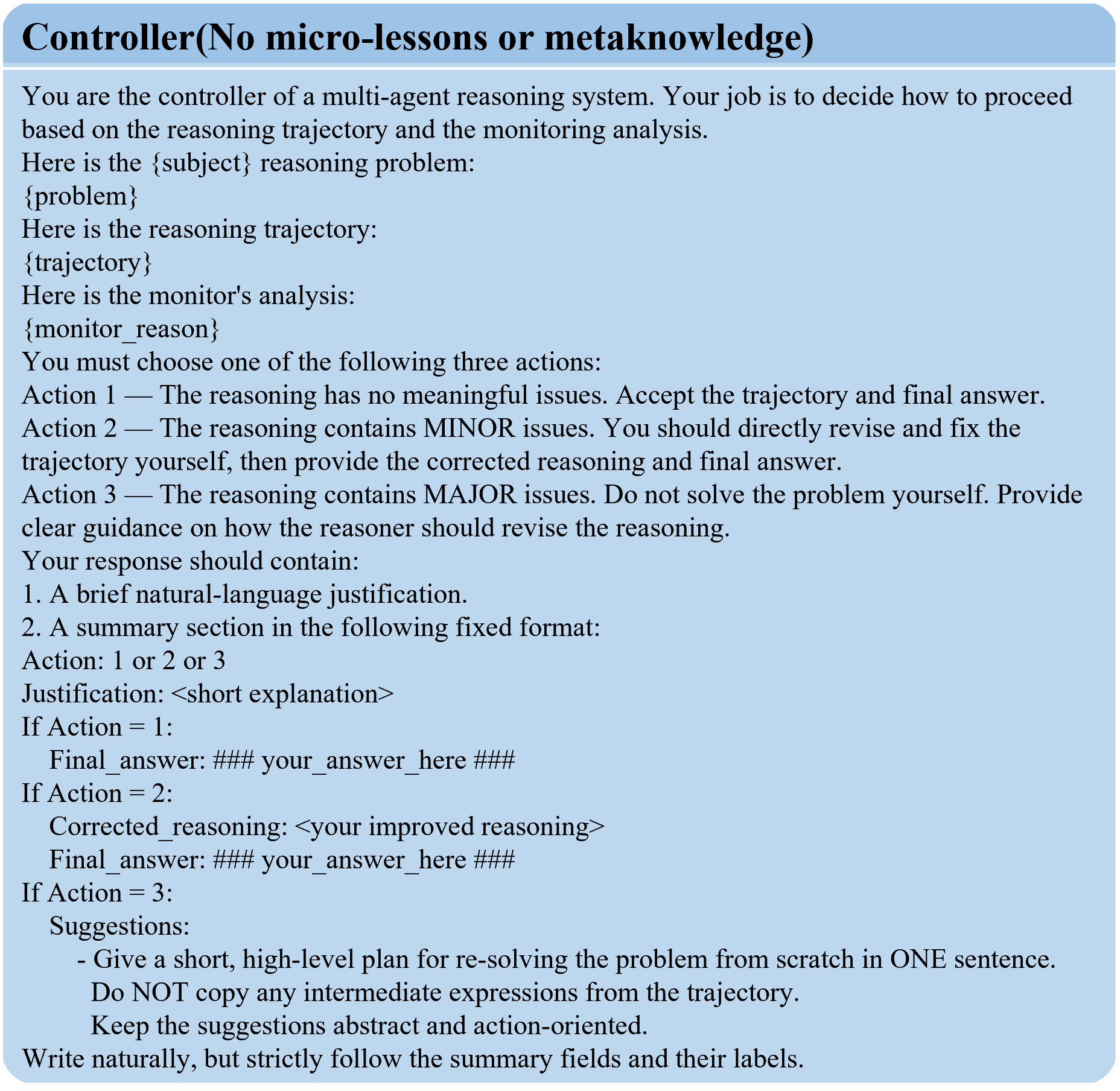}
{Controller prompt (cold start; no micro-lessons or meta-knowledge).}
{fig:prompt_controller_coldstart}

\promptfig{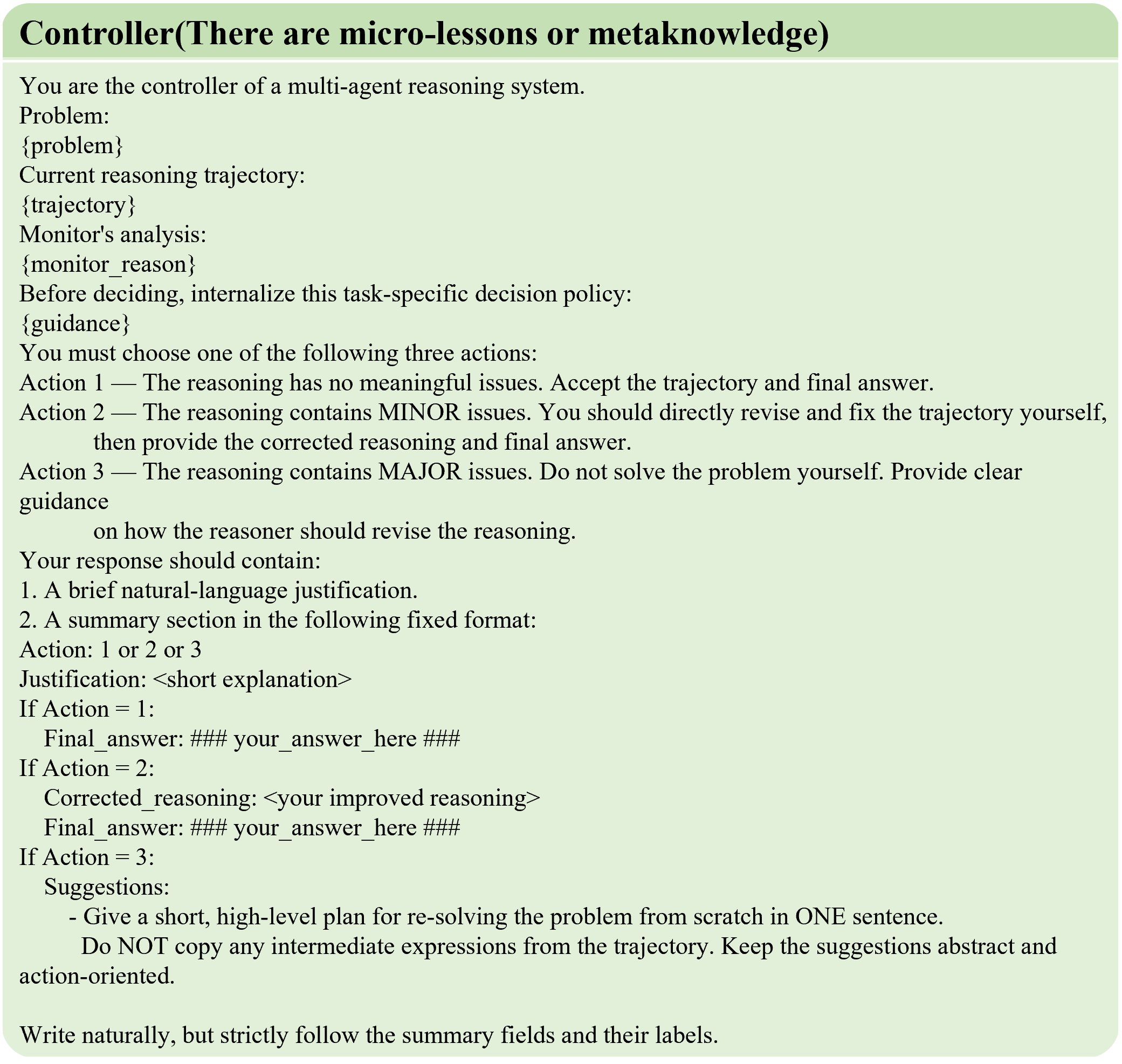}
{Controller prompt (with retrieved micro-lessons / meta-knowledge decision policy).}
{fig:prompt_controller_guided}

\promptfig{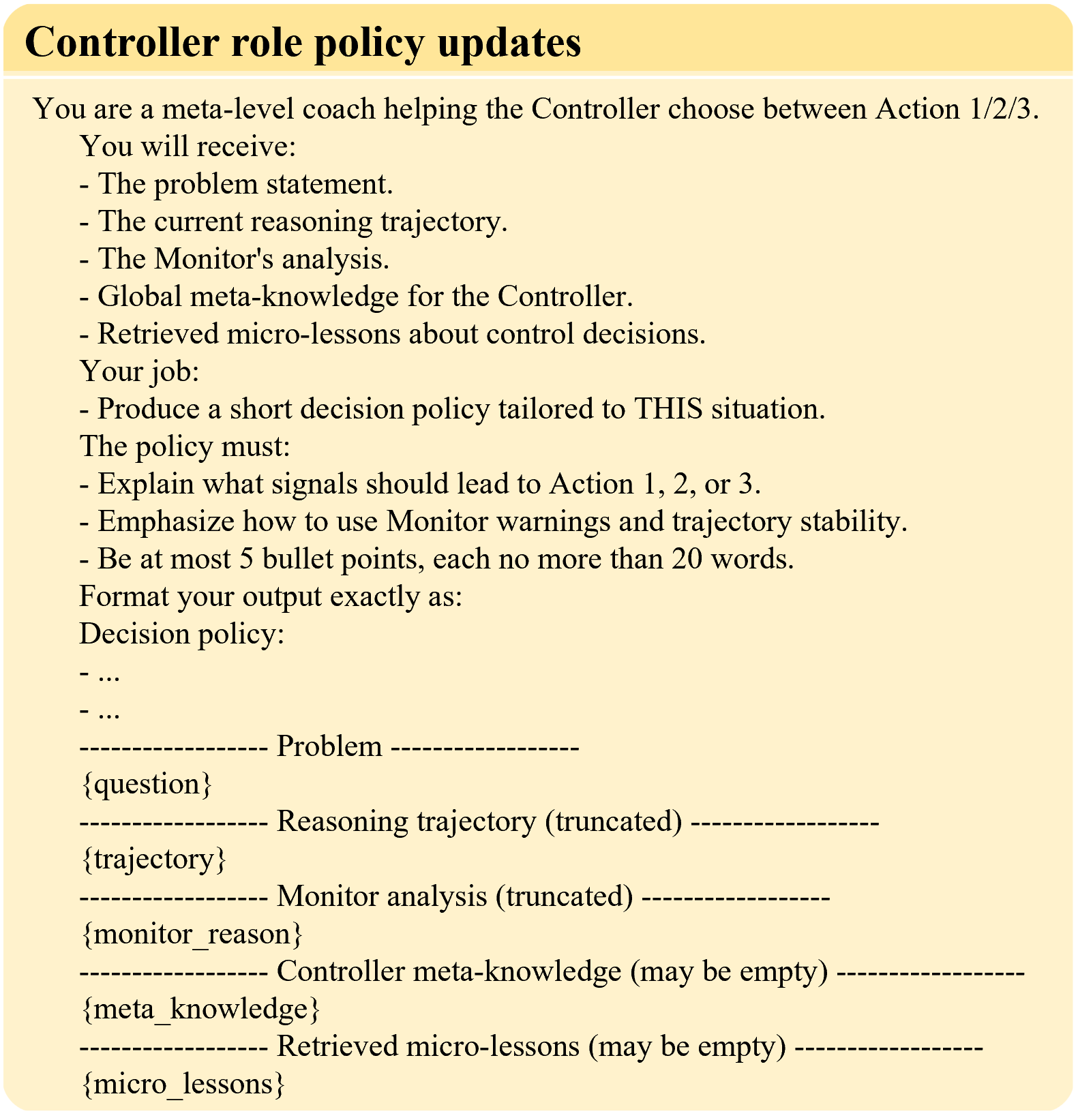}
{Controller role policy update prompt (meta-level coaching to produce a concise decision policy).}
{fig:prompt_controller_policy_update}

\promptfig{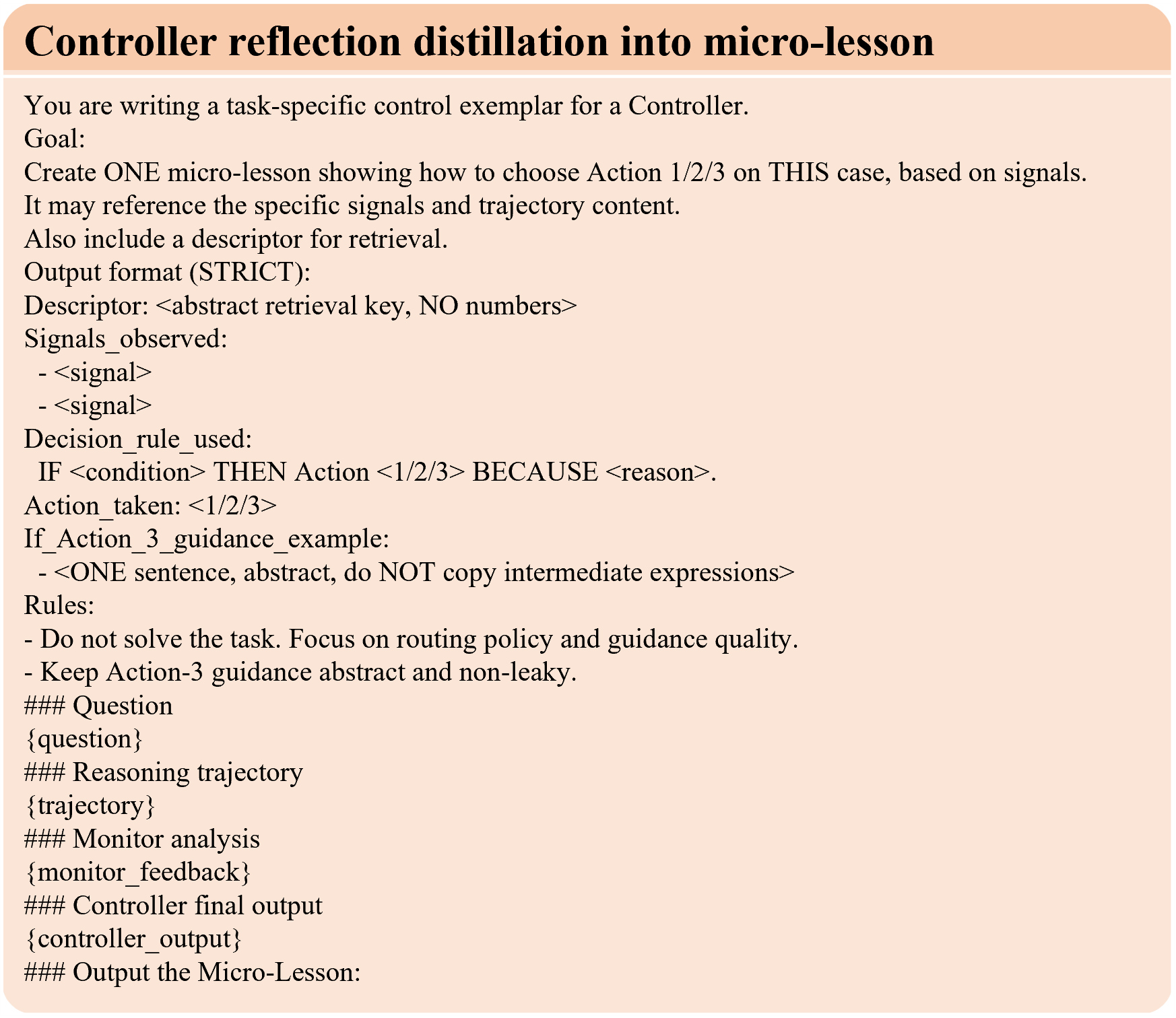}
{Controller reflection distillation prompt for micro-lesson (control exemplar mapping signals to Action choices).}
{fig:prompt_controller_distill}

\promptfig{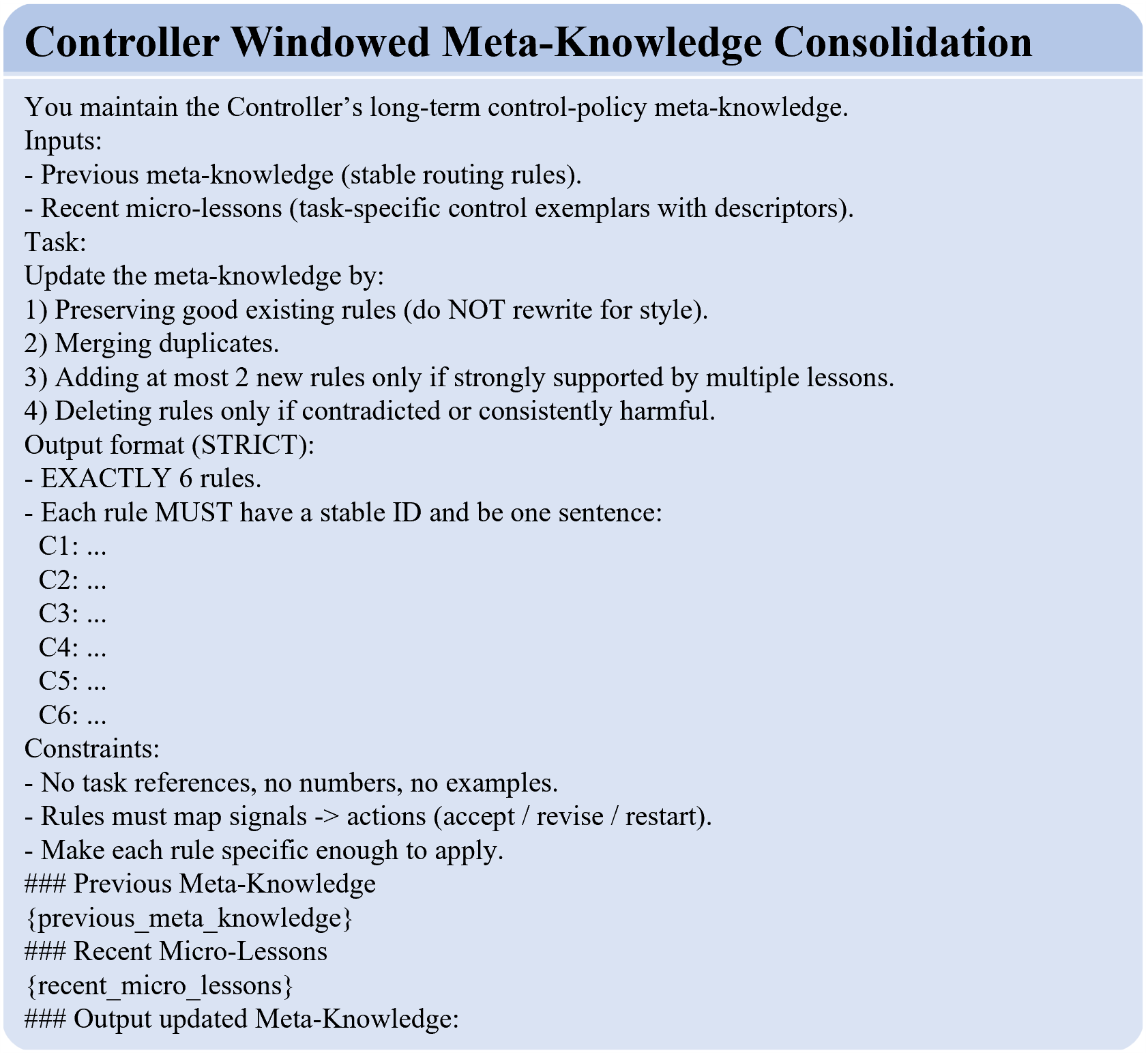}
{Controller prompt for windowed meta-knowledge consolidation (update long-term control-policy rules from recent micro-lessons).}
{fig:prompt_controller_consolidate}

\endgroup

\end{document}